\documentclass[conference]{IEEEtran}
\usepackage{amssymb,amsfonts}
\usepackage{graphicx}
\usepackage{textcomp}
\usepackage{ctable}
\usepackage{cite}
\usepackage{soul}
\usepackage{algorithm}
\usepackage{algpseudocode}
\floatname{algorithm}{Algorithm}

\usepackage{amsmath}

\newcommand{\vecv}{\vec{v}\@ifnextchar{^}{\,}{}}
\newcommand{\vect}{\vec{t}\@ifnextchar{^}{\,}{}}

\usepackage{multirow}
\usepackage{array}
\usepackage{diagbox}
\usepackage{xspace}

\newcommand{\et}{\textit{et al.\xspace}}

\newcommand{\sysname}{\texttt{Forsaken}\xspace}

\newcommand{\protocol}[1]{\texttt{#1}}

\usepackage{subfigure}
\usepackage{epstopdf}
\usepackage{fancyhdr}
\usepackage{boxedminipage}
\usepackage{wrapfig}

\usepackage{ntheorem}
\theorembodyfont{\itshape}
\theoremheaderfont{\scshape}
\theoremseparator{.}
\newtheorem{definition}{Definition}

\makeatletter
\def\ALG@special@indent{%
    \ifdim\ALG@thistlm=0pt\relax
        \hskip-\leftmargin
    \else
        \hskip\ALG@thistlm
    \fi
}
\newcommand{\algsubtitle}[1]{\item[]\noindent\ALG@special@indent \textbf{#1}}
\makeatother

\makeatletter
\newcommand{\multiline}[1]{%
  \begin{tabularx}{\dimexpr\linewidth-\ALG@thistlm}[t]{@{}X@{}}
    #1
  \end{tabularx}
}
\makeatother

\usepackage[font=small,labelfont=bf]{caption}

\definecolor{reviewA}{HTML}{FF0000}
\definecolor{reviewB}{HTML}{0000FF}
\definecolor{reviewC}{HTML}{008000}
\definecolor{reviewD}{HTML}{FF00FF}
\definecolor{reviewE}{HTML}{800000}
\definecolor{common}{HTML}{F0FFF0}
\definecolor{ly}{HTML}{008080}

\usepackage{longtable}
\newcommand\revision[1]{{#1}}

\newcommand*{\textlabel}[2]{%
  \edef\@currentlabel{#1}
  \phantomsection
  #1\label{#2}
}\usepackage{hyperref}

\newcommand{\Paragraph}[1]{~\vspace*{-0.8\baselineskip}\\{\bf #1}}

\begin{document}
\title{Learn to Forget: Machine Unlearning via Neuron Masking}
%

\makeatletter
\newcommand{\linebreakand}{%
  \end{@IEEEauthorhalign}
  \hfill\mbox{}\par
  \mbox{}\hfill\begin{@IEEEauthorhalign}
}
\makeatother

\author{
\IEEEauthorblockN{Yang Liu}
\IEEEauthorblockA{bcds2018@foxmail.com}
\and
\IEEEauthorblockN{Zhuo Ma}
\IEEEauthorblockA{mazhuo@mail.xidian.edu.cn}
\and
\IEEEauthorblockN{Ximeng Liu}
\IEEEauthorblockA{snbnix@gmail.com}
\and
\IEEEauthorblockN{Jian Liu}
\IEEEauthorblockA{jian.liu@eecs.berkeley.edu}
\linebreakand
\IEEEauthorblockN{Zhongyuan Jiang}
\IEEEauthorblockA{zyjiang@xidian.edu.cn}
\and
\IEEEauthorblockN{JianFeng Ma}
\IEEEauthorblockA{jfma@mail.xidian.edu.cn}
\and
\IEEEauthorblockN{Philip Yu}
\IEEEauthorblockA{psyu@uic.edu}
\and
\IEEEauthorblockN{Kui Ren}
\IEEEauthorblockA{kuiren@zju.edu.cnm}
}


\maketitle              

\begin{abstract}
Nowadays, machine learning models, especially neural networks, become prevalent in many real-world applications. 
These models are trained based on a one-way trip from user data: as long as users contribute their data, there is no way to withdraw; and it is well-known that a neural network memorizes its training data.
This contradicts the ``right to be forgotten'' clause of GDPR, potentially leading to law violations.  
To this end, 
{\em machine unlearning} becomes a popular research topic, 
which  allows users to eliminate memorization of their private data from a trained machine learning model.

In this paper, we propose the first uniform metric called forgetting rate to measure the effectiveness of a machine unlearning method. It is based on the concept of membership inference and describes the transformation rate of the eliminated data from ``memorized'' to ``unknown'' after conducting unlearning. 
We also propose a novel unlearning   method called \sysname. It is superior to previous work in either utility or efficiency (when achieving the same forgetting rate).
We benchmark \sysname with eight standard datasets to evaluate its performance. 
The experimental results show that it can achieve more than 90\% forgetting rate on average and only causeless than 5\%  accuracy loss.
\end{abstract}


%
%
%


\section{Introduction}
\label{sec_introduction}
Nowadays, machine learning models, especially neural networks, become prevalent in many real-world applications including medical diagnosis, credit-risk assessment, autopilot and so on. 
These models are trained from user data that may contain sensitive information. 
Recent research, like federated learning~\cite{bonawitz2017practical, tran2019federated} and cryptography-based machine learning~\cite{mohassel2018aby3, agrawal2019quotient}, enables the training  process to be done in a way without seeing the training data.
However, the trained model itself may still ``memorize'' the training data~\cite{feldman2020does}, and there is no way for users to withdraw.
Recently released data protection regulations, e.g., the California Consumer Privacy Act~\cite{harding2019understanding} and the General Data Protection Regulation in the European Union~\cite{voigt2017eu}, clearly state that users should have the right to withdraw their private data.

To this end, {\em machine unlearning} (MU) becomes a popular research topic.
It  allows users to eliminate memorization of their private data from a trained machine learning model.
\revision{
    \textlabel{At first glance, differential privacy (DP)}{rA:1}~\cite{geyer2017differentially} can naturally achieve machine unlearning. 
    It guarantees that by looking at the model, one cannot tell whether a sample is in the training data or not.
    However, DP focuses on protecting the privacy of all samples and the protection is only to some extent.  
    More specifically, DP ensures a subtle bound on the contribution of each sample to the final model, but the contribution cannot be constrained to zero; otherwise, the model would learn nothing from the training data.
    In contrast, the aim of MU is to cancel the contribution of a {\em target} sample {\em completely}.
    Therefore, MU is orthogonal to DP.
}

Existing unlearning methods can be roughly classified into two categories: summation-based~\cite{cao2015towards} and retraining-based~\cite{bourtoule2019machine, neel2021descent}.
Summation-based unlearning trains a model based on a small number of {\em summations}, each of which is the sum of some efficiently computable transformation (e.g., gradient descent) of the training samples~\cite{cao2015towards}.
To forget a sample, one can simply subtract that sample from its corresponding summations and update the model.
This method works well for non-adaptive machine learning models (later training does not depend on earlier training), such as Na\"ive Bayes~\cite{saritas2019performance} and C4.5~\cite{quinlan1996bagging}.
However, for adaptive models such as neural networks, subtracting a sample from a summation can easily cause excessive unlearning of unrelated memorization, hence significantly decrease the utility.

As its name implies, retraining-based unlearning retrains the model after removing the samples to be forgotten~\cite{ginart2019making}.  
Given that retraining from scratch incurs an overhead that is usually unaffordable (it may take several days to retrain a model),  
Bourtoule \et~\cite{bourtoule2019machine} propose SISA, which divides the training set into slices and trains a model via incrementally learning, s.t. each intermediate model (after one slice is added) is recorded.
To forget a sample, retraining starts from the first intermediate model that contains the contribution of that sample.
However, this method is simply trading storage (for recording the intermediate models) for retraining time, instead of truly reducing the overhead of retraining.

\Paragraph{Our contributions.}
In this paper, we propose a novel unlearning method called \sysname.
Compared with summation-based unlearning, \sysname is more friendly to adaptive models such as neural networks, 
introducing a much smaller utility loss.
Compared with retraining-based unlearning, \sysname has a much more efficient unlearning phase in both storage and time usage.
The core idea of \sysname is {\em neuron masking}.
We introduce a mask gradient generator that continuously generates mask gradients, and apply them to the neurons of the neural network and stimulate them to unlearn the memorization of the given samples. 
Based on the state of the updated model, the mask gradient generator adjusts the magnitude of the mask gradients to avoid unexpected unlearning.

Especially, \sysname can be applied to unlearn any training data, including out-of-distribution (OOD) and in-distribution (ID) data, but focus more on OOD data unlearning.
This is because the OOD but sensitive data inadvertently uploaded by users can form unintended memorization, which is hard to avoid and can increase the possibility of the adversary to extract private user information~\cite{carlini2019secret}.

Furthermore, we propose a uniform metric called \textit{forgetting rate} to evaluate the effectiveness of an unlearning method.
It is based on the membership oracle~\cite{shokri2017membership, salem2018ml}, which is to infer whether a given sample is a member of the training set.
It describes the transformation rate of the eliminated data from ``memorized'' to ``unknown'' after conducting memorization elimination.
To the best of our knowledge, this is the first indicator that can be directly used for machine unlearning evaluation.

The contributions of this paper are summarized below.
\vspace{-0.2cm}
\begin{itemize}
    \item 
    We propose {\bf the first uniform metric} called forgetting rate to  measure the effectiveness of a machine unlearning method. (Section~\ref{sec_approach})
    \vspace{-0.2cm}
    
    \item 
    We propose {\bf a novel machine unlearning method} called \sysname.
    It is superior to previous work in either utility or efficiency (when achieving the same forgetting rate).
    (Section~\ref{sec_sysname})
    \vspace{-0.2cm}
    
    \item
    We benchmark \sysname with eight standard datasets to evaluate its performance.
    The experimental results show that it can achieve {\bf more than 90\% forgetting rate} on average and only cause {\bf less than 5\% accuracy loss}. (Section~\ref{sec_experiments})
\end{itemize}
\vspace{-0.4cm}

\section{Background}
\label{sec_motivation}
This section briefly introduces the necessary technical background for understanding this paper.

\subsection{Machine unlearning} 
For machine unlearning, prior work mainly explores two ways, summation-based MU~\cite{cao2015towards} and retraining-based MU~\cite{bourtoule2019machine, neel2021descent}.

The former one is implemented based on statistical query learning~\cite{kearns1998efficient}.
Denote the summation of training sample transformation at the $i_{th}$ iteration as $G_{i} = \sum_{x\in D} g_{i}(x)$, where $g_{i}(x)$ is the transformation of a specific sample $x$, $D$ is the training set. 
According to summation-based MU, a model trained for $k$ iterations can be expressed as: $f_{k} = Learn(G_{1}, G_{2}, ..., G_{k})$.
When there are samples $D'$ required to be unlearned, summation-based MU computes the equation below.
\begin{equation}\label{eq_naitve_eliminate}
\begin{aligned}
    f_{k}' = 
    Learn(&G_{1} - G'_{1},
    G_{2} - G'_{2}, ...,
    G_{k} - G'_{k}),\\
    &G'_{i} = \sum_{x\in D'}g_{i}(x),
\end{aligned}
\end{equation}
where $f_{k}'$ is the updated model whose memorization about $D'$ is forgotten.
Summation-based MU works well for non-adaptive machine learning models.
As for adaptive neural networks, a slight parameter change at one iteration can make all the subsequent training to be deviated, which can further cause a considerable decrease in model performance.

The latter one introduces the ensemble learning technique to implement machine unlearning.
Specifically, retraining-based MU divides the whole training set into multiple shards, and each shard is further partitioned into multiple slices, which are used to train a constituent model slice by slice.
During the training process, every intermediate model updated after one slice is added is recorded.
When machine unlearning is invoked, retraining is applied to the first intermediate model trained without the slice containing the forgotten data.
Although such a method is more efficient than native full retraining, its ensemble learning mechanism still needs massive computation and storage resources.
Also, due to the partition of the training set, the performance of constituent models cannot be guaranteed. 

Inspired by the above illustration, a formal definition to describe machine unlearning is given as follows.

Foremost, a solid fact about machine learning is that no matter taking which kind of training method, it is the training set that determines what a model memorizes.
If the memorization of a sample is unlearned, the most direct consequence is that the model no longer biases the posterior probability~\cite{salem2019updates} of forgotten data to follow member data posterior distribution.
Such a statement also explains why the recently proposed membership inference attack~\cite{shokri2017membership, rahman2018membership} can infer whether a sample is involved in the training.
Based on the principle, we propose Definition~\ref{def_k_elimination} that formally rules what the state of a machine learning model is supposed to be after conducting machine unlearning.

\vspace{-0.1cm}
\begin{definition}[Machine Unlearning]\label{def_k_elimination}
Given a model $f$, an ideal membership oracle $\psi$, a training set $D$ and an unlearning set $D'\subset D$, machine unlearning is perfectly performed if there exists a function $\varphi$ that can output $f^* = \varphi(f)$ where for each $x\in D'$, we have $\psi(f(x)) = true$ and $\psi(f^*(x)) = false$, and meanwhile, for each $x\in D/D'$, $f^*(x) = f(x)$.
\end{definition}
\vspace{-0.1cm}

\revision{Moreover, given a learning task, the training data can always be divided into two types of data, namely OOD data and ID data.
Correspondingly, machine unlearning also involves OOD data unlearning and ID data unlearning.
Here, we give the formal definition~\mbox{\cite{hsu2020generalized}} of these two types of data.}
\begin{definition}[\revision{OOD and ID Data}]\label{def_ood}
\revision{
Given a learning task, assume the desired distribution of the task to be $\chi$.
For the sample $x\sim \chi$, we say that $x$ is in-distribution (ID), otherwise, it is out-of-distribution (OOD). 
}
\end{definition}

\subsection{Membership oracle}
A {\em membership oracle} (or {\em  membership inference attack}) answers queries of whether a certain sample is in the training set of a machine learning model.
It can be implemented based on one or multiple shadow models~\cite{shokri2017membership, salem2018ml, nasr2018machine}.
More specifically, an attacker collects a shadow dataset that is in the same distribution as the training set of the target model;
and trains a shadow model based on the shadow dataset.
Then, the attacker feeds some members and non-members of the shadow model's training set to the shadow model, and gets a set of confidence vectors.
Based on these confidence vectors, the attacker trains a binary classifier that can decide whether a confidence vector output by a model is from an input that belongs to the model's training set.
Then, given a sample's confidence vector output by the target model, the binary classifier can be used to infer whether this sample is a member of the target model's training set.

In this paper, we utilize the membership oracle as a tool to evaluate the effectiveness of a machine unlearning method. 
Notice that existing membership oracles can only be used in classification tasks.
Therefore, the machine learning models mentioned in this paper are all classifiers by default.

\section{Forgetting Rate}
\label{sec_approach}
In this section, we propose the first uniform metric called {\em forgetting rate} that can measure the effectiveness of a machine unlearning method.
Given a set of samples $D$ to be forgotten, the forgetting rate (FR) of a machine unlearning method is defined as follows:
\begin{equation}\label{eq_forgetting_rate}
\begin{gathered}
    \mathrm{FR} = \frac{\mathrm{AF} - \mathrm{BF}}{\mathrm{BT}}
\end{gathered}
\end{equation}
\begin{itemize}
    \item AF: the number of samples in $D$ that are predicted to be \textbf{false} by a membership oracle \textbf{after} machine unlearning.
    \item BF: the number of samples in $D$ that are predicted to be \textbf{false} by a membership oracle \textbf{before} machine unlearning.
    \item BT: the number of samples in $D$ that are predicted to be \textbf{true} by a membership oracle \textbf{before} machine unlearning.
\end{itemize}

In this way, FR gives an intuitive measurement of the rate of samples that are changed from member to non-member after unlearning.
An effective unlearning method should at least achieve AF $>$ BF on the condition that BT $>0$, 
otherwise, unlearning is reversed or meaningless. 
When FR $= 100\%$, AF $=$ BT + BF $= |D|$, which means that all memorized samples have been successfully unlearned.
On the other hand, when FR $= 0\%$, AF $=$ BF, which means no memorized samples have been unlearned.
We further remark that the membership oracle can be any membership inference algorithm.

\section{\sysname}
\label{sec_sysname}

In this section, we formally introduce \sysname, a novel machine unlearning method that is superior to previous work in either utility or efficiency.
We first give an intuition for designing \sysname and then provide more details.

\subsection{Intuition}

The core idea of \sysname is masking the neurons of the target model with gradients (called {\em mask gradients}) that are trained to eliminate the memorization of some training samples from the target model.
This idea was inspired from the theory of ``active forgetting''~\cite{davis2017biology} in Neurology.
Different from the hysteresis of natural forgetting (a.k.a ``passive forgetting''), active forgetting is vigorous and it can eliminate all traces and engram cells for a given memory.
To achieve active forgetting, the forgetting cells produce a kind of special dopamine, which serves as the forgetting signal that stimulates remodeling of the engram cells and accelerates memorization elimination.
Based on the feedback of the engram cells, the forgetting cells can adjust dopamine production to avoid unexpected forgetting.

In \sysname, a {\em mask gradient generator} $\mathcal{G}$ plays the role of forgetting cells.
To eliminate the memorization of some training samples,
\sysname continuously invokes $\mathcal{G}$ to produce mask gradients, which is similar to dopamine.
The mask gradients are then applied to the neurons of the neural network and stimulate them to eliminate the memorization of the given samples.
Based on the state of the updated model, $\mathcal{G}$ adjusts the magnitude of the mask gradients to avoid catastrophic forgetting.
Fig.~\ref{fig_dummy_gradient} visualizes this process.

\begin{figure}[ht!]
\centering
\includegraphics[scale=0.68]{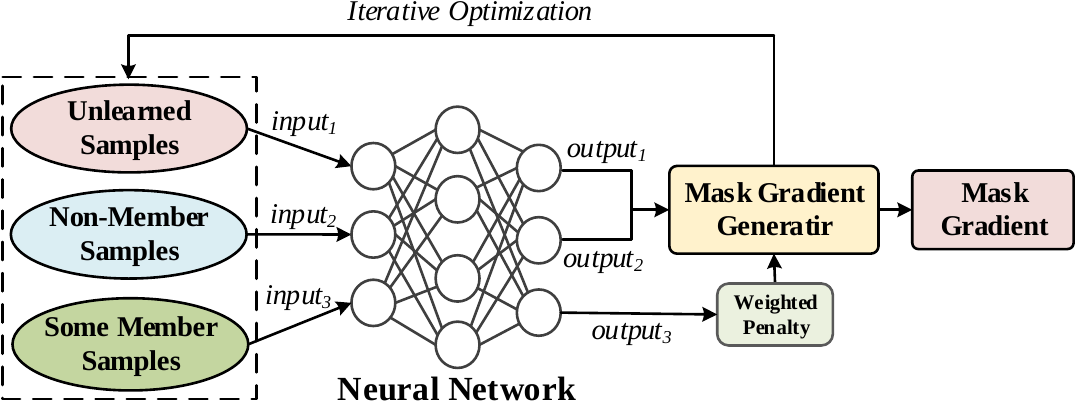}
\caption{The high-level workflow of mask gradient generator.}
\label{fig_dummy_gradient}
\end{figure}

\vspace{0.1cm}
\noindent
\textbf{Improvements.}
Compared with the existing methods, the mask gradient based \sysname mainly achieves the following improvements.

\vspace{-0.2cm}
\begin{enumerate}
    \item \sysname can be applied to all kinds of the existing neural networks, no matter deep or shallow architecture, and meanwhile, does not introduce obvious accuracy loss on the original model.
    
    \vspace{-0.2cm}
    \item \sysname does not break the standard model training procedure (avoid retraining and uploading data in a predefined order) or change the original model architecture.
\end{enumerate}

\subsection{\sysname in detail}\label{sub_generator}
Algorithm~\ref{pro_mem_generator} shows the pseudocode for \sysname and Table~\ref{table_notation_2} lists the notations.
At the $t$-th iteration, 
we first generate the mask gradients $\mu_{t}$ with the mask gradient generator $\mathcal{G}$ (line 3),
and then update the parameters $\theta_{t-1}$ with $\mu_{t}$ (line 4).
Next, we feed each sample to be forgotten into the updated model $f_{\theta_{t}}$ and obtain \revision{\textlabel{the prediction results}{rE7:3}:  $\Upsilon = \{y_i | y_i \gets f_{\theta_{t}}(x_i), x_i\in D\}$ (line 5).
For a given task to classify any input into one of $p$ classes, the output $y_i\in \Upsilon$ is limited to be a confidence vector with fixed $p$ dimensions.}
In the end of the $t$-th iteration, we optimize $\mathcal{G}$ with $\Upsilon$ by minimizing $\mathcal{L}_{forgetting}$, which is the distance between $\Upsilon$ (i.e., the posteriors of $D$) and $\mathcal{P}$ (i.e., the posterior distribution of non-member data) (line 6).
During the optimization procedure, the original parameters of the target model are fixed.
There are two challenges for optimizing $\mathcal{G}$:
1) defining its loss function $\mathcal{L}_{forgetting}$;
2) determining a proper posterior distribution $\mathcal{P}$ for non-member data.
Next, we explain how we address these two challenges in greater details.

\renewcommand\arraystretch{1.2}
\begin{table}[!htbp]
\centering
\footnotesize
\caption{Notation Table}
\begin{tabular}{p{1.1cm}<\centering| l}

\specialrule{.1em}{.05em}{.05em} 
Notations      & Descriptions \\

\hline
$\mu_t$      &  The mask gradients optimized after $t$ iterations.  \\

$\xi$        &  The forgetting coefficient of the mask gradient. \\

$D$         &  The set of samples to be forgotten. \\

$\Upsilon$   &  The posteriors of $D$. \\

$\theta_0$   &  The initial trainable parameters of the target model. \\

$\mathcal{P}$ & The posterior distribution of non-member data. \\

$T$ &  The maximum training iterations. \\

$\lambda$ &  the penalty coefficient. \\
\specialrule{.1em}{.05em}{.05em} 
\end{tabular}
\label{table_notation_2}
\end{table}
\vspace{-0.2cm}

\begin{algorithm}[ht!]
  \caption{\sysname in a nutshell}
  \label{pro_mem_generator}
  \begin{algorithmic}[1]
    \Require 
    A target model $f_{\theta_0}$ and its trainable parameters $\theta_0$;
    A set of samples $D$ to be forgotten; 
    \revision{$T$ is the maximum number of training iterations.}
    
    \Ensure
      A final model $f_{\theta_T}$.
    
    \State Initialize the mask gradient $\mu_0$

    \For{$t\gets 1$ to $T$}
        \State 
        $\mu_{t}\gets \mathcal{G}(\theta_{t-1}, \Upsilon)$.
        \State 
        $\theta_{t}\gets \theta_{t-1} - \xi \cdot \mu_{t}$.
        
        \State 
        $\Upsilon = \{y_i | y_i \gets f_{\theta_{t}}(x_i), 	\forall  x_i\in D\}$.
        
        \State Optimize $\mathcal{G}$ by minimizing $\mathcal{L}_{forgetting}$.
    \EndFor    
    
    \State \revision{Delete $\mathcal{G}$, and $D$ from the original dataset.}
    \State \textbf{Return} $f_{\theta_T}$.
  \end{algorithmic}
\end{algorithm}

\subsubsection{Loss function of $\mathcal{G}$} 

The goal of the mask gradient generator $\mathcal{G}$ is to find a particular perturbation over $\theta$ to minimize the distance between $\Upsilon$ (i.e., the posteriors of $D$) and $\mathcal{P}$ (i.e., the posterior distribution of non-member data):

\vspace{-0.05cm}
\begin{equation}\label{eq_g_op_problem}
\begin{aligned}
     \quad &\mathcal{L}(\Upsilon, \mathcal{P})\\
\end{aligned}
\end{equation}

To resolve this optimization problem, we need to first quantify the distance between member and non-member posteriors.
We borrow experience from another trendy topic of machine learning security, adversarial example (AE)~\cite{goodfellow2016nips}.
To generate an AE,  one needs to estimate the distance between the posteriors of an AE and a specific legal output for a given target model.
A common distance function used in AE generation is Kullback-Leibler divergence (KL-divergence)~\cite{goodfellow2016nips}.
We adopt this function in $\mathcal{G}$.

We also need to avoid ``over-unlearning'', which could potentially lower the accuracy of the target model. 
To this end, we adopt a penalty item, L$_1$ norm regularization item~\cite{donoho2006most}, 
which is also commonly used in normal neural network training to avoid overtraining.
The reason for choosing L$_1$ norm, not L$_p$ norm, $p \geq 2$, is that L$_1$ norm is smoother and the penalty of L$_p$ norm can block the convergence of \sysname.

Put it altogether, we can rewrite Eq.~\ref{eq_g_op_problem} as follows.
\vspace{-0.1cm}
\begin{equation}\label{eq_g_op_problem_improved}
\begin{aligned}
    \quad &\mathcal{L}_{KL}(\Upsilon, \mathcal{P}) + \lambda\cdot |\!| \mu |\!|_1\\
\end{aligned}
\end{equation}

where $\mathcal{L}_{KL}$ represents \textit{KL} divergence loss function, $\lambda$ is the penalty coefficient, $|\!| \mu |\!|_1$ is the L$_1$ norm regularization item and $\mu$ is the mask gradient.
\revision{
\textlabel{We say that the above optimization problem can always work.}{rE1:1}
If the KL-divergence of $P$ and the posteriors of $D$ are already close (i.e., have a small KL-divergence), that means the target model does {\em not} remember $D$ in the very beginning. 
Then, there is no need of unlearning $D$.
Otherwise, any optimizer can be used to solve the above problem.
}
Moreover, we do not use the common first-order gradient optimizer as $\mathcal{G}$, like SGD and adaptive moment estimation (Adam)~\cite{bock2019proof}, but the second-order gradient optimizer, L-BFGS~\cite{zhu1997algorithm}, which is slower than the first-order optimizer but converge more steadily.

\revision{
    \textlabel{Further, in our evaluation, we find that}{cr4:1} the above loss function still suffers from a dilemma of balancing the forgetting rate and accuracy drops.
    In particular, the traditional penalty item treats all parameter changes equally.
    Sometimes, such a penalty mechanism leads to some updates that can cause dramatic performance drops to be not blocked, and some changes that are useful to ``forgetting'' but not affects model performance to be overblocked.
    To resolve the problem, we propose a dynamic penalty mechanism as follows.
}
\vspace{-0.1cm}
\begin{equation}\label{eq_g_op_problem_improved_improved}
\begin{aligned}
   &\mathcal{L}_{KL}(\Upsilon, \mathcal{P}) + \lambda\cdot\omega\cdot |\!| \mu |\!|_1
\end{aligned}
\end{equation}
\revision{where $\omega$ is designed to maintain a subtle bound of penalty over parameter changes.
When $\omega = 1$, Eq.~\mbox{\ref{eq_g_op_problem_improved_improved}} and Eq.~\mbox{\ref{eq_g_op_problem_improved}} are identical.
$\omega$ can be computed based on Eq.~\mbox{\ref{eq_penalty_omega}}.
}
\begin{equation}\label{eq_penalty_omega}
    \omega = \frac{1}{|D_0|} \sum_{x\in D_0} |\frac{\partial \mathcal{L}_{cross}(x)}{\partial \theta(d)}|,
\end{equation}
\revision{where $\theta(d)$ specifies the $d_{th}$ dimension of parameters and $D_0$ is a very small set of training data (less than 1\%) that should not be forgotten.
It can be observed that $\omega$ increases positively with the partial gradients of $D_0$.
Note that the partial gradients potentially indicate the performance degradation level of the target model on normal training data.
Therefore, if the change of $\theta_0$ results in higher performance degradation, we assign more penalty to mitigate such catastrophic forgetting, and vice versa.
In Section~\mbox{\ref{sub_evaluation_factor}}, we implement both Eq.~\mbox{\ref{eq_g_op_problem_improved}} and \mbox{Eq.~\ref{eq_g_op_problem_improved_improved}} to justify the improvement of above modification.
}

\subsubsection{Non-member data} 

Recall that the goal of the mask gradient generator $\mathcal{G}$ is to minimize the distance between $\Upsilon$ (i.e., the posteriors of $D$) and $\mathcal{P}$ (i.e., the posterior distribution of non-member data).
$\Upsilon$ can be obtained from the prediction results after feeding the samples to the model.
We still need to estimate $\mathcal{P}$.

\revision{
    Roughly, the training data can be classified into two categories, OOD data and ID data.
    Here, we first discuss how to derive the desired posterior distribution of non-member OOD data, and then, expand it to ID data.
}

The intuition for estimating non-member OOD data is that the model no longer strongly predicts a non-member OOD sample to be in any specific category.
Then, based on Shannon entropy theory~\cite{lin1991divergence}, we define an ideal non-member OOD posterior distribution as follows.

\vspace{-0.1cm}
\begin{definition}[Ideal Non-member OOD posterior]\label{def_forgotten}
Based on Shannon's entropy theory~\cite{shannon2001mathematical}, for \revision{a given neural network $f_{\theta}$ with $p$-dimensions output}, we say that an OOD sample $x$ is ideally non-member when it reaches maximum entropy, i.e., its posterior $(y_1, ..., y_p) = f_{\theta}(x)$ satisfying $y_1=y_2 =... =y_p =\frac{1}{p}$.    
\end{definition}
\vspace{-0.1cm}

However, in real-world scenarios, such ideal posterior distribution is rarely seen.
Even for the randomly generated meaningless data that has never occurred in the model training process, their posteriors are always biased towards some specific dimensions due to the generality of neural networks~\cite{zhang2016understanding}.
\revision{
    Therefore, instead of using the ideally uniform posterior distribution, we sample the posterior distribution of some real-world OOD data to serve as $\mathcal{P}$, defined in Definition~\ref{def_forgotten_normal}.
}
Since the target distribution tends to be similar to the normal real-world non-member OOD samples, the difficulty of the adversary to identify the forgotten data is increased, and the catastrophic forgetting can be mitigated.

\begin{definition}[Real-World Non-member OOD]\label{def_forgotten_normal}
For a neural network $f_{\theta}$ with $p$-dimensions output, we say that an OOD sample $x$ is real-world non-member if its posteriors $(y_1, y_2, ..., y_p) = f_{\theta}(x)$ tends to be $(\overline{z}_1, \overline{z}_2, ..., \overline{z}_p) = f_{\theta}(X_n)$, where $X_n$ is a set of non-member OOD samples and $\overline{z}_i$ is the posterior distribution of $X_n$ over dimension $i$.
\end{definition}

In more details, the sampling of real-world non-member OOD distribution can be implemented by leveraging some publicly available data that are irrelevant to the learning task.
For example, assume the learning task is \textit{0-9} digital number recognition.
We can first collect digital images about \textit{a}-\textit{z} characters to serve as the non-member OOD data, and label them by querying the target model.
Then, the posteriors of the non-member OOD data with the same class are averaged to form $\mathcal{P}$.
\revision{
    Inspired by the above discussion, we can simply expand \sysname to the ID data unlearning scenario by replacing the non-member OOD data with testing data.
}

\begin{definition}[Real-World Non-member ID]\label{def_forgotten_id_data}
For a neural network $f_{\theta}$ with $p$-dimensions output, we say that an ID sample $x$ is real-world non-member if its posteriors $(y_1, y_2, ..., y_p) = f_{\theta}(x)$ tends to be $(\overline{z}_1, \overline{z}_2, ..., \overline{z}_p) = f_{\theta}(X_n)$, where $X_n$ is a set of non-member ID samples (testing samples) and $\overline{z}_i$ is the posterior distribution of $X_n$ over dimension $i$.
\end{definition}

\revision{
    Furthermore, compared to OOD data, ID data are more highly related to the learning task, and thus, non-member ID posterior is usually closer to its corresponding member posterior.
    The fact always leads to more accuracy reduction and slower convergence of ID data unlearning than OOD data unlearning.
    In Section~\ref{sec_experiments}, comprehensive experiments are conducted to validate such a statement.
}

\subsection{How to apply \sysname in applications}\label{sub_apply_me}
Given the design details of \sysname, we present Algorithm~\ref{pro_mem_elimination} (\protocol{MacForget}) to explain how to apply in applications.

\begin{algorithm}[ht!]
  \caption{Federated Learning with Machine Unlearning (\protocol{MacForget})}
  \label{pro_mem_elimination}
  \begin{algorithmic}[1]
    \Require
    the user $u_0$, 
    the local dataset $D_0$ of $u_0$ with the size $n_0$;
    the current neural network $f_{\theta_{k}}$;
    the learning rate $\eta$;
    the loss function $\mathcal{L}(\cdot)$;
    the maximum iterations for machine unlearning $T$;
    
    \Ensure
      The updated model $f_{\theta_{k + 1}}$.
      
    \State Each user $u_0$ who is involved in the $k_{th}$ iteration of training does the following steps.

    \If{$u_0$ chooses to conduct machine unlearning} 
        \State $u_0$ first samples the posterior distribution $\mathcal{P}$ of non-member data based on $f_{\theta_{k}}$.
        
        \State {$u_0$ then computes mask gradient $\mu\gets$ \sysname$(f_{\theta_k}, \theta_k, D_0, T, \mathcal{P})$ and sends $\mu' \gets \frac{1}{\eta}\cdot n_0\cdot \mu$ to $\mathcal{S}$.}
    \Else    
        \State {$u_0$ computes the normal gradient $\nabla \mathcal{L}(D_0, \theta_k)$ and then, sends $\nabla\mathcal{L}(D_0, \theta_k)$ to $\mathcal{S}$.}
    \EndIf

    \State $\mathcal{S}$ accumulates the (mask) gradients collected from all users to the current model.
    
    \State \textbf{Return} the updated model $f_{\theta_{k+1}}$.
  \end{algorithmic}
\end{algorithm}

Take the federated learning scenario~\cite{yang2019federated, liu2020federated} as an example, in which a server $\mathcal{S}$ collects user gradients to train a neural network.
In this case, the user who wants to withdraw some previously uploaded data can conduct \sysname and upload the mask gradient to $\mathcal{S}$ instead of normal gradient (line 4).
\revision{Since the mask gradient has identical size and similar value range to the normal gradient, $\mathcal{S}$ can treat mask gradients in the same way as normal gradients (line 8), the correctness of which is given below.
\begin{equation}\label{eq_acc_dummy_gradient}
\begin{aligned}
    \theta_k - \eta\cdot \frac{1}{n_0} \cdot \mu = \theta_k - \eta\cdot \frac{1}{n_0} \cdot \frac{1}{\eta}\cdot n_0\cdot\mu = \theta_k - \mu,
\end{aligned}
\end{equation}
where $\theta_k$ is the parameters of the target model at $k_{th}$ iteration, $\eta$ is the learning rate of target neural network, $n_0$ is the size of a normal training batch, and $\mu$ is the mask gradient output by \sysname.
Based on the algorithm, users can efficiently complete its private data unlearning in only one iteration and make it hard to infer which user conducts machine unlearning.}

\section{Evaluation}
\label{sec_experiments}
In this section, comprehensive experiments are conducted to evaluate the effectiveness and efficiency of \sysname on both OOD and ID data.

\subsection{Experimental Setup}

\vspace{0.1cm}
\noindent
\textbf{Dataset.}
We mainly implement \sysname on eight standard datasets (details are attached in Appendix~\ref{sec_appendix_dataset}.), namely 
CIFAR10,
CIFAR-100, 
IMDB,
News,
Reuters,
STL10,
TinyImage,
Customer.
The datasets cover three different types of learning tasks, including image classification, sentiment analysis and text categorization.

\vspace{0.1cm}
\noindent
\textbf{OOD Data.}
\revision{
    \textlabel{We leverage}{rA:3} three common ways~\cite{hendrycks2018deep} to simulate OOD data.
}

\begin{enumerate}
    \item \textit{C10.S.} 
    The OOD data are from the same learning task but mislabeled by users, e.g., CIFAR10 and STL-10.
    We randomly select parts of data from STL-10 and mislabel them with another label to serve as OOD data.
    
    \item \textit{C10.T., C100.T. and I.C.} 
    The OOD data are from completely different learning tasks.
    For example, we can collect skirt images from TinyImages, label them as birds and insert them into CIFAR10, which does not contain skirt images, to serve as OOD data.
    
    \item \textit{News (15-5) and Reuters (35-16).}
    In this case, the training set is split into two parts based on labels.
    One is used for target model training. 
    The other is used as the OOD set.
    For instance, we can select samples from News whose labels are between 0-14 as the target set.
    The others are deployed as OOD data.
\end{enumerate}
\revision{
    Note that all unlearning OOD data are randomly selected with a mix of labels in our evaluation.
    The default unlearning size is 200.
    Moreover, 1000 samples are randomly chosen from the remaining unselected OOD samples to compute $\mathcal{P}$.
    Table~\ref{table_ood_data} summarizes the usage of the datasets in the experiments.
}

\renewcommand\arraystretch{1.2}
\begin{table}[!htbp]
\centering
\footnotesize
\caption{OOD data simulation}
\begin{tabular}{|c|c|c|}

\hline
Abbreviation & Target Dataset & OOD Dataset \\

\hline
C10.S. & CIFAR10 & STL-10 \\

\hline
C10.T. & CIFAR10 & TinyImage \\

\hline
C100.T. & CIFAR100 & TinyImage \\

\hline
I.C. & IMDB & Customer Review \\

\hline
Reuters (35-11) & Reuters (35) & Reuters (11) \\

\hline
News (15-5) & News (15) & News (5) \\


\hline
\end{tabular}
\label{table_ood_data}
\end{table}

\vspace{0.1cm}
\noindent
\revision{
\textbf{ID Data.}
The unlearning ID data are randomly selected from the training set.
Besides, we randomly select 1000 samples from the testing set to compute the desired non-member ID distribution defined in Definition~\ref{def_forgotten_id_data}.}

\vspace{0.1cm}
\noindent
\textbf{Membership Oracle.}
To compute FR, we train a black-box membership oracle for each dataset according to the shadow model based method~\cite{salem2018ml}.
The performance of each membership oracle is given in Appendix~\ref{sec_appendix_membership_inference}.
Concretely, we randomly partition each dataset into two halves, which are used to train the target and the shadow model, respectively.
Two models have the same architecture and hyperparameters.
Then, we train an XGBoost~\cite{chen2016xgboost} model (the best performance model in our experiments) to serve as the membership oracle.
\revision{In Section~\ref{sec_appendix_white_box}, we further discuss the performance of \sysname with the white-box inference oracle.}

\vspace{0.1cm}
\noindent
\textbf{Neural Networks.}
Four different neural networks are involved in processing the target datasets.
The detailed neural network architectures are given in Appendix~\ref{sec_appendix_network}.

\vspace{0.1cm}
\noindent
\textbf{Other Setting.}
The experiments are simulated on a workstation equipped with Tesla V100 GPU and 16GB VRAM.
\revision{The experiment results are averaged over ten trials and processed in a single thread.}

\subsection{Performance Comparison}\label{sub_sec_performance_comp}
To evaluate the performance of machine unlearning, three indicators should be considered: 
1) the number of samples that are memorized during training, i.e., $BT$ in Eq.~\ref{eq_forgetting_rate}; 
2) the rate of samples that are successfully unlearned, i.e., $FR$; 
3) \revision{the performance drops compared with the original model.}
Indicator 1 suggests whether the memorization of unlearning data is formed.
Indicator 2 evaluates the effectiveness of machine unlearning algorithms.
Indicator 3 indicates the side-effect of machine unlearning.

\renewcommand\arraystretch{1.2}
\begin{table*}[t!]
\centering
\footnotesize
\caption{\revision{{Forgetting rate of \sysname on different datasets for OOD data unlearning}}}

\resizebox{0.8\textwidth}{!}{%
\begin{tabular}{c|c|c|c|c|c|c|c}

\toprule
\multirow{2}{*}{\textbf{Dataset}}  & \multirow{2}{1.2cm}{\centering \textbf{Parameter}\\ \textbf{Size}} & \multirow{2}{*}{\textbf{BT (BF)}} & \multicolumn{5}{c}{\textbf{Forgetting Rate (Average / Variance)}}\\

\cline{4-8}
&  &   & \multicolumn{1}{c|}{\textbf{Full Retraining}} & \multicolumn{1}{c|}{\textbf{\sysname}} & \multicolumn{1}{c|}{\textbf{SMU~\cite{cao2015towards}}} & \multicolumn{1}{c|}{\textbf{SISA~\cite{bourtoule2019machine}}} & \multicolumn{1}{c}{\textbf{SISA-DP~\cite{neel2021descent}}}  \\


\hline
C10.S. & 14.77M & 168 (32)        & 93.45 / 0.34\%  & 97.62 / 0.44\%   & 86.91 / 4.89\%  & 91.67 / 0.79\% & 94.05 / 1.21\% \\

\hline
C10.T. & 14.77M & 176 (24)        & 94.89 / 0.28\%  & 98.29 / 0.69\%   & 93.75 / 5.94\%  & 96.03 / 0.63\% & 97.16 / 0.73\% \\

\hline
C100.T. & 14.77M & 117 (83)       & 96.58 / 0.41\%  & 95.73 / 1.32\%   & 71.79 / 4.13\%  & 86.32 / 2.89\% & 89.74 / 1.59\%  \\
\hline
I.C. & 2.62M & 195 (5)            & 94.36 / 0.16\%  & 98.46 / 0.79\%   & 43.59 / 6.33\%  & 95.89 / 0.41\% & 96.41 / 1.14\% \\

\hline
Reuters (35-11) & 5.26M & 192 (8) & 94.79 / 0.13\%  & 96.35 / 0.82\%   & 81.25 / 2.51\%  & 95.32 / 0.76\% & 93.75 / 0.87\% \\

\hline
News (15-5) & 0.25M & 162 (38)    & 95.06 / 0.23\%  & 97.53 / 0.59\%   & 86.93 / 4.47\%  & 92.26 / 1.09\% & 94.44 / 1.81\% \\
\bottomrule




\end{tabular}
}
\label{table_forgetting_rate}
\end{table*}

\renewcommand\arraystretch{1.2}
\begin{table*}[!t]
\centering
\footnotesize
\caption{\revision{{Forgetting rate of \sysname on different datasets for ID data unlearning}}}

\resizebox{0.78\textwidth}{!}{%
\begin{tabular}{c|c|c|c|c|c|c|c}

\toprule
\multirow{2}{*}{\textbf{Dataset}}  & \multirow{2}{1.2cm}{\centering \textbf{Parameter}\\ \textbf{Size}} & \multirow{2}{*}{\textbf{BT (BF)}} & \multicolumn{5}{c}{\textbf{Forgetting Rate (Average / Variance)}}\\

\cline{4-8}
&  &   & \multicolumn{1}{c|}{\textbf{Full Retraining}} & \multicolumn{1}{c|}{\textbf{\sysname}} & \multicolumn{1}{c|}{\textbf{SMU~\cite{cao2015towards}}} & \multicolumn{1}{c|}{\textbf{SISA~\cite{bourtoule2019machine}}} & \multicolumn{1}{c}{\textbf{SISA-DP~\cite{neel2021descent}}}  \\


\hline
CIFAR10 & 14.77M & 167 (33)     & 85.32 / 0.09\%   & 88.63 / 2.29\%   & 85.03 / 5.77\%   & 86.23 / 0.41\%   & 88.02 / 0.43\% \\

\hline
CIFAR100 & 14.77M & 191 (9)     & 94.24 / 0.04\%   & 87.96 / 1.63\%   & 79.06 / 5.05\%   & 84.29 / 1.52\%   & 88.48 / 1.46\%  \\
\hline
IMDB & 2.62M & 151 (49)         & 84.77 / 0.11\%   & 94.04 / 2.19\%   & 62.25 / 6.13\%   & 87.42 / 0.71\%   & 90.72 / 0.81\% \\

\hline
Reuters  & 5.26M & 160 (40)     & 83.13 / 0.09\%   & 86.25 / 1.29\%   & 69.38 / 5.33\%   & 81.25 / 0.29\%   & 84.38 / 0.76\% \\

\hline
News & 0.25M & 172 (28)         & 85.47 / 0.71\%   & 90.12 / 1.84\%   & 86.86 / 6.67\%   & 84.31 / 0.37\%   & 87.21 / 0.84\% \\
\bottomrule




\end{tabular}
}
\label{table_forgetting_rate_id}
\end{table*}

\revision{Table~\ref{table_forgetting_rate} and Table~\ref{table_forgetting_rate_id} summarize the former two indicators and compare \sysname with other four machine unlearning methods on both OOD and ID data unlearning.
Here, native \textit{full retraining} \textlabel{without the unlearning data}{cr2:1} serves as the basic baseline.}
To implement summation-based MU~\cite{cao2015towards} (SMU), we record all gradients of the unlearning data in each epoch, and then, conduct one more epoch of training to subtract them from the target model.
\revision{As for the retraining-based MU proposed in~\mbox{\cite{bourtoule2019machine}} (SISA), \textlabel{we set the number of shards to}{rC3:1} be $10$, a very low level of partition in applications (ten users in federated learning).
Each shard is then divided into 100 slices, which are used to train a constituent model slice by slice.
The output of SISA is decided by majority voting.}
The posterior of SISA is the mean value of the posteriors for the constituent models that are winners in the voting process.
The hyperparameters of each constituent model in SISA are the same as \sysname.
For fairness, the membership oracle used for SISA is the same as \sysname and SMU, not trained by the ensemble of multiple shadow constituent models. 
Such a configuration can also show that roughly applying ensemble learning to machine unlearning may cause unexpected performance loss, such as weaker memorization on training data, lower accuracy and more training time.
\revision{
    SISA-DP~\cite{neel2021descent} is implemented similar to SISA.
    The only difference is that SISA-DP introduces more tricks to ensure model performance and includes a DP based model publishing function.
    The differential noise budget $\epsilon$ is set to be $16$ as suggested by~\mbox{\cite{leino2020stolen}}, which is often used in applications and can maintain a good balance between accuracy drops and data privacy.
}

From the results, the first observation is that although OOD data are irrelevant to the original learning task, most of them are still memorized by the target model. 
Besides, most of the involved OOD data are correctly classified into the correct classes and do not obviously affect the performance of the target model (shown in Table~\ref{table_neural_network_performance}, Appendix~\ref{sec_appendix_dataset}).
The experiments prove that unintended memorization caused by OOD data is a common phenomenon in real applications.
Further, we present detailed experiment analysis as follows.

\vspace{0.1cm}
\noindent
\textbf{Forgetting Rate.}
\revision{
    For both OOD and ID data, \sysname \textlabel{generally achieves close to}{cr2:2}, or even higher, $FR$ than full retraining, and the size of the training set or the mask gradient does not strongly affect the performance of \mbox{\sysname}.
    Relatively, SMU has the worst performance due to its non-directional forcible unlearning mechanism.
    Intuitively, without consideration of practicality, the most effective method to implement machine unlearning is full retraining.
    However, full retraining fails to achieve 100\% $FR$ in our experiments.
    The reason is that the generality of neural networks makes the retrained model to be still capable of identifying some unlearning data with high confidence, especially in the ID data scenario.
    In contrast, \mbox{\sysname} focuses on all memorization related to the unlearning data, no matter general and specified.
    Therefore, \mbox{\sysname} outperforms retraining based methods on $FR$ sometimes but the tradeoff is a little more accuracy drops.
    As for SISA-DP, due to the differential noises, it achieves higher $FR$ than SISA but suffers from more accuracy drops.
    Here, an interesting discovery is that on the model protected with a low level of noise, membership inference attains similar performance to the non-defensive model, which corresponds to the statement of~\mbox{\cite{leino2020stolen}}.
    Moreover, compared to OOD data unlearning, ID data unlearning achieves lower $FR$.
    This is due to the high relevance of ID data to the learning task, which makes ID data harder to be unlearned and easily cause more accuracy drops..
}


\begin{figure}[htbp]
\centering
\subfigure[Test accuracy after machine unlearning.]{
    \begin{minipage}[t]{0.45\linewidth}
    \centering
    \includegraphics[scale=0.24]{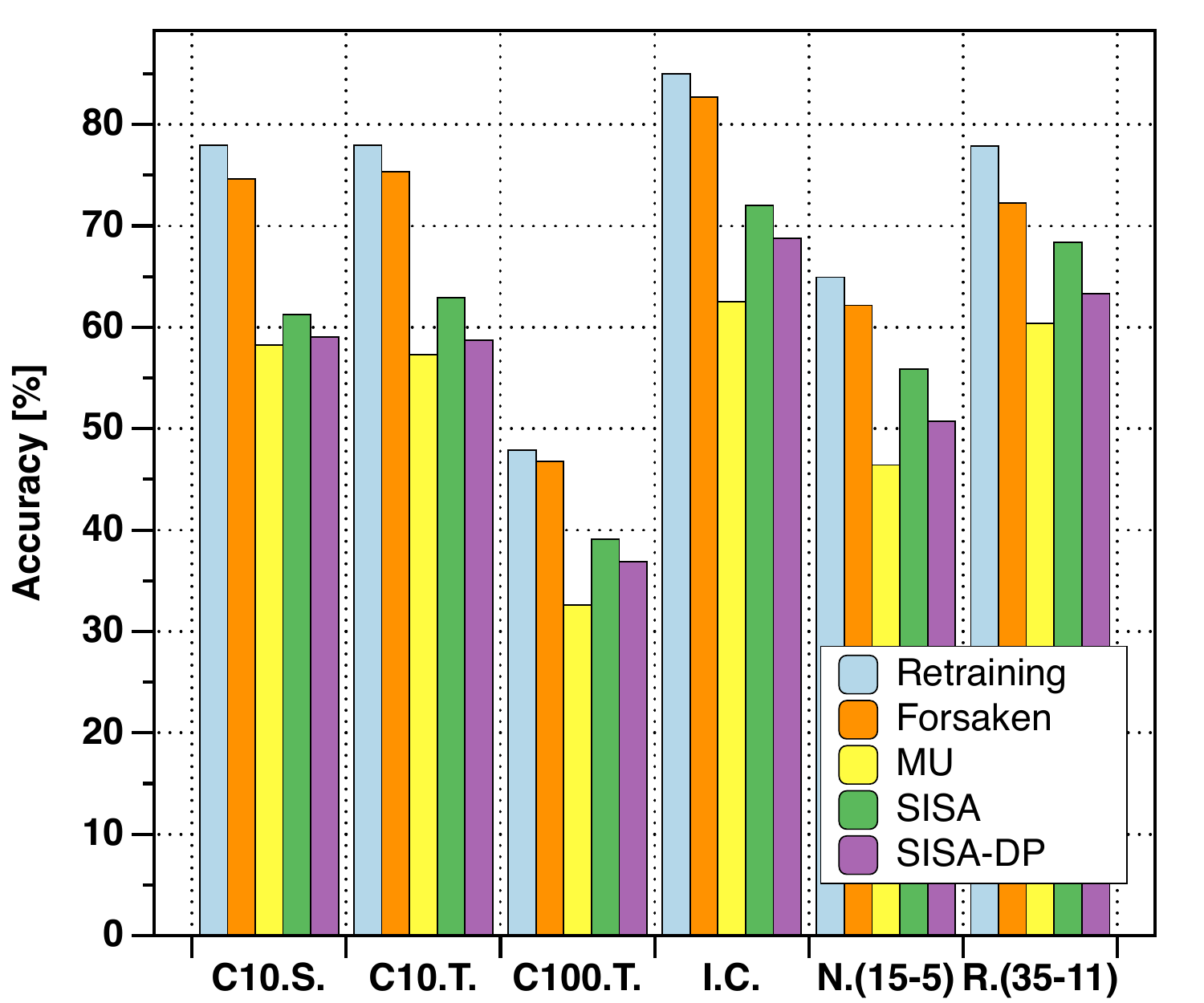}
    \label{fig_test_acc}
    \end{minipage}
}%
\hfill
\subfigure[Test loss after machine unlearning.]{
    \begin{minipage}[t]{0.45\linewidth}
    \centering
    \includegraphics[scale=0.24]{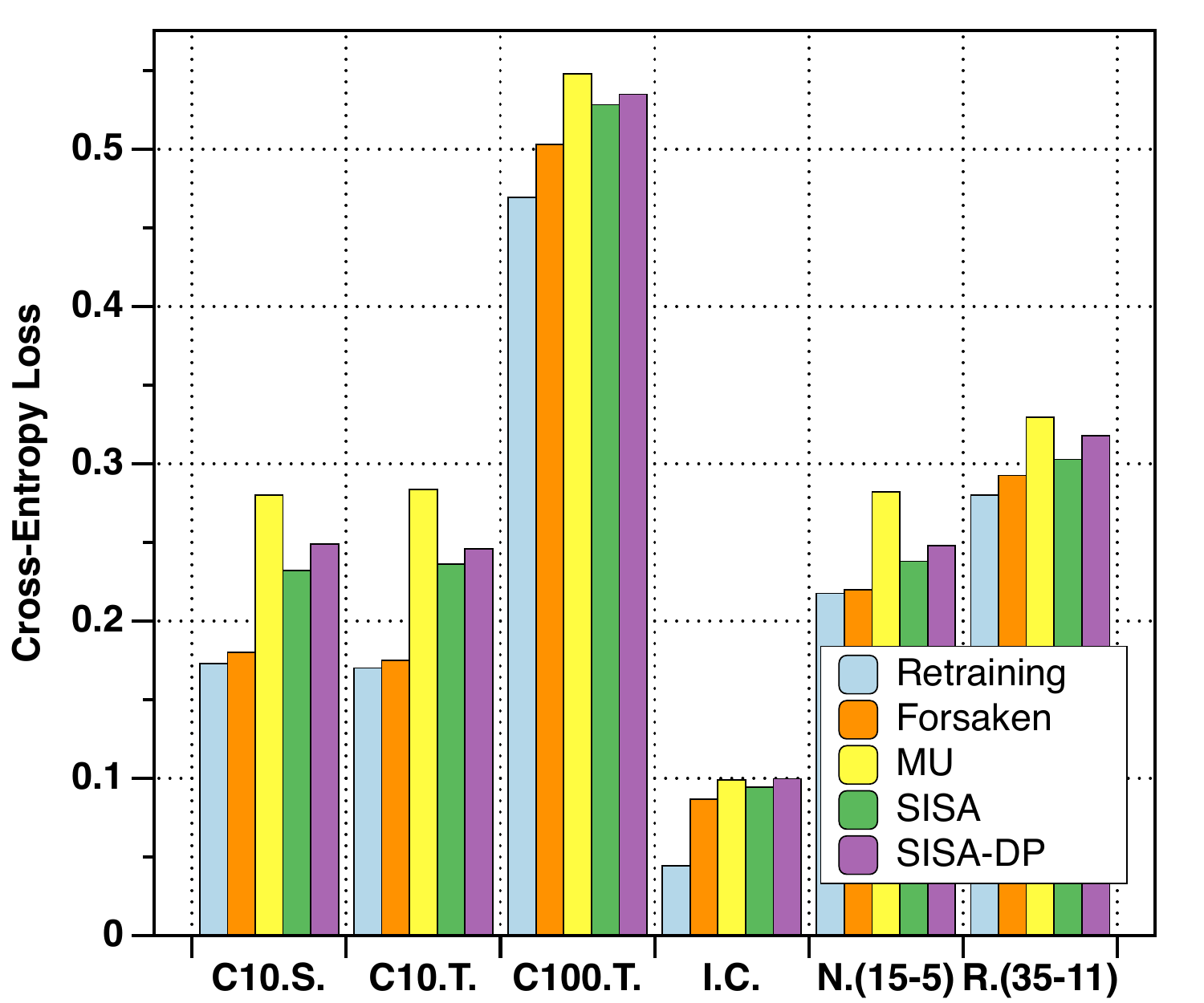}
    \label{fig_test_loss}
    \end{minipage}
}%
\caption{\revision{Performance change of the target model over the testing set for OOD data unlearning.}}
\label{fig_performance_loss}
\end{figure}

\vspace{-0.2cm}
\begin{figure}[htbp]
\centering
\subfigure[Test accuracy after machine unlearning.]{
    \begin{minipage}[t]{0.45\linewidth}
    \centering
    \includegraphics[scale=0.24]{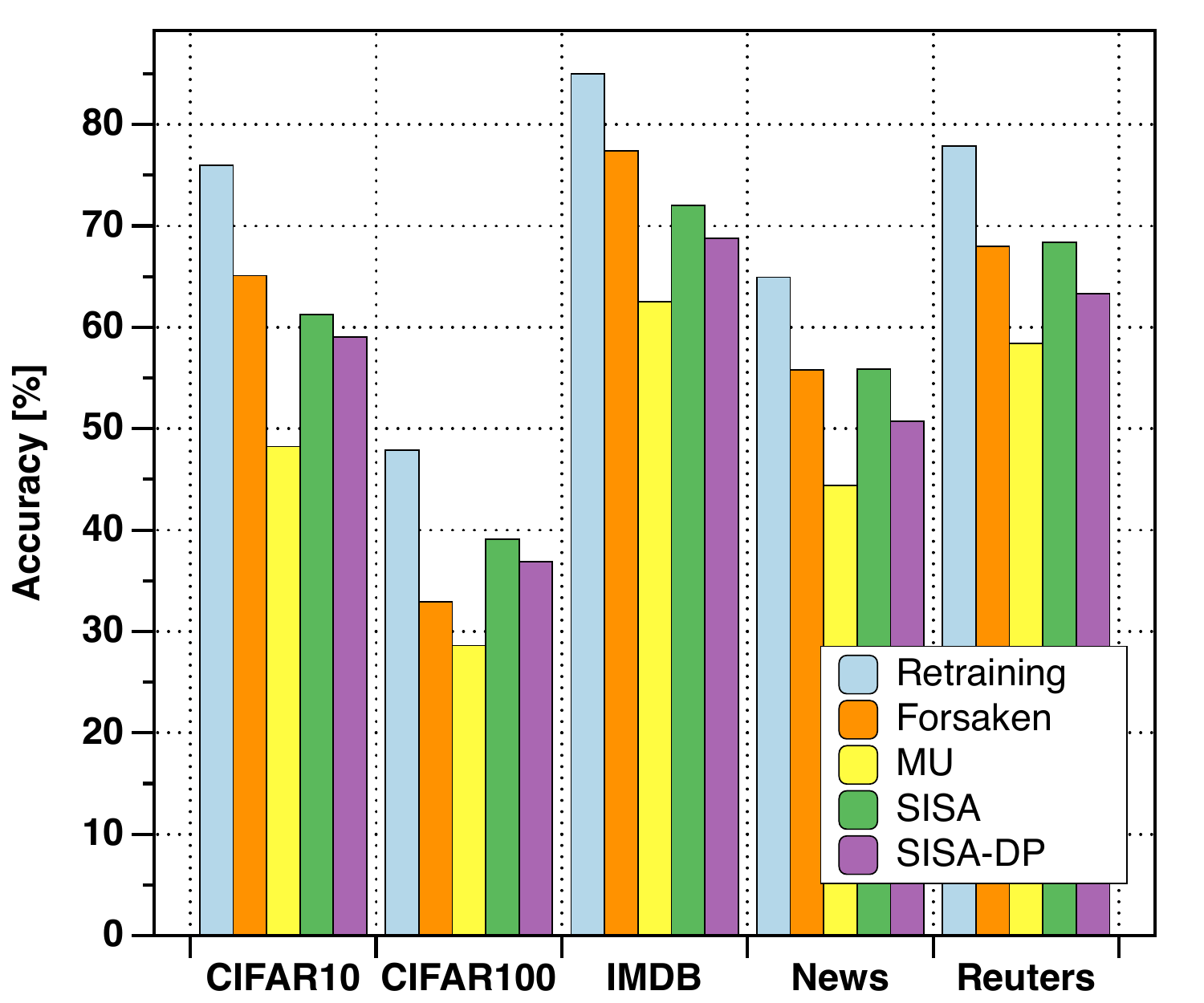}
    \label{fig_test_acc_id}
    \end{minipage}
}%
\hfill
\subfigure[Test loss after machine unlearning.]{
    \begin{minipage}[t]{0.45\linewidth}
    \centering
    \includegraphics[scale=0.24]{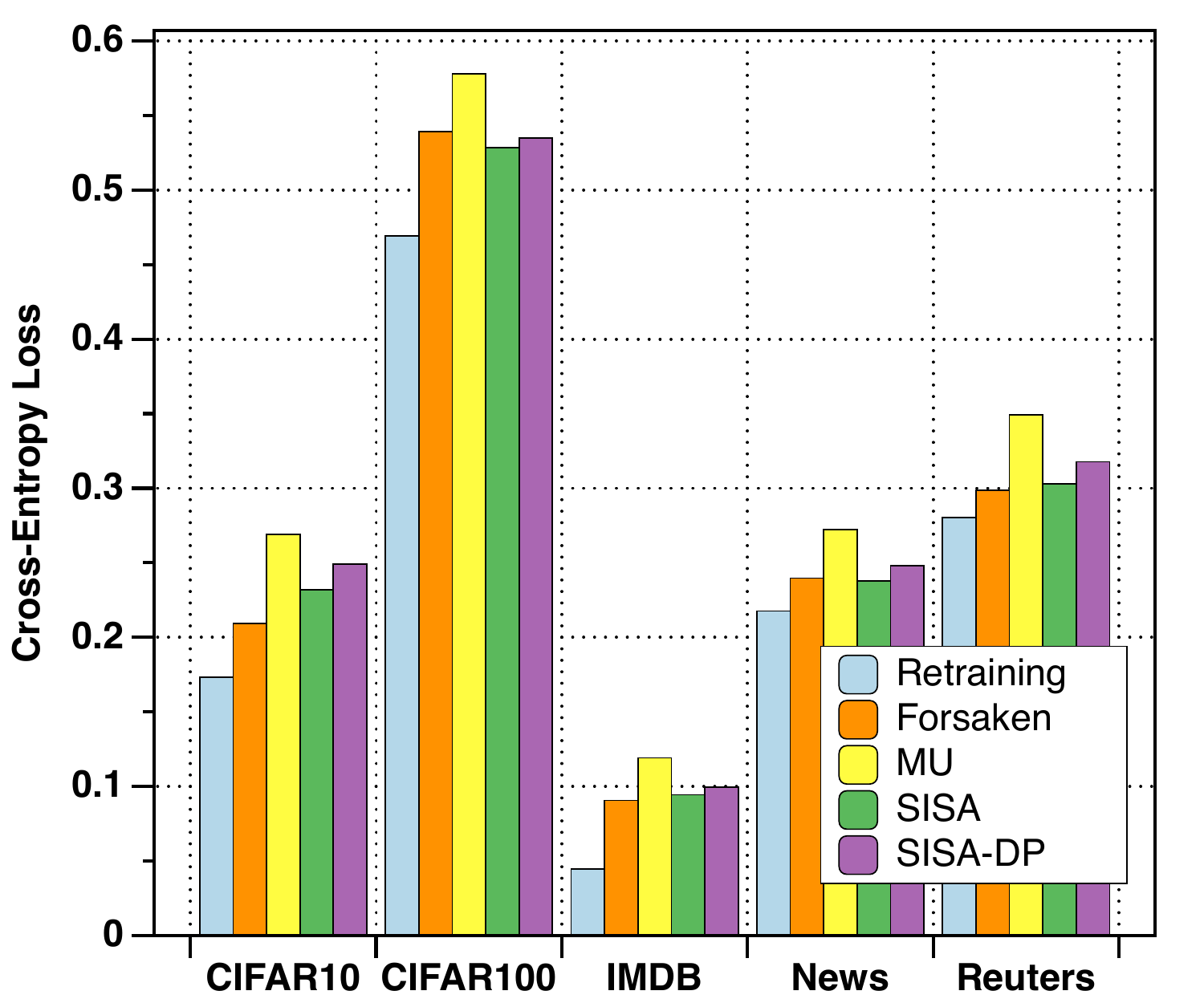}
    \label{fig_test_loss_id}
    \end{minipage}
}%
\caption{\revision{Performance change of the target model over the testing set for ID data unlearning.}}
\label{fig_performance_loss_id}
\end{figure}

\vspace{0.1cm}
\noindent
\textbf{Performance Drops.}
\revision{Then, we focus on the model performance change of the target model before and after conducting OOD and ID data unlearning, shown in Fig.~\ref{fig_performance_loss} and Fig.~\ref{fig_performance_loss_id}, respectively.
From the results, the model performances after conducting SISA and SISA-DP are much lower than the original normal model, which reflects its weaker memorization over the training set caused by data partition.
Conversely, except full retraining, the performance drop of \mbox{\sysname} is the lowest due to its subtle remodeling of the target model.
Moreover, as mentioned before, the performance drop caused by ID data unlearning is higher than OOD data for \sysname.}

\begin{figure}[t!]
\centering
\subfigure[The change of $FR$ with different unlearning sizes for C10.T.]{
\begin{minipage}[t]{0.45\linewidth}
\centering
\includegraphics[scale=0.24]{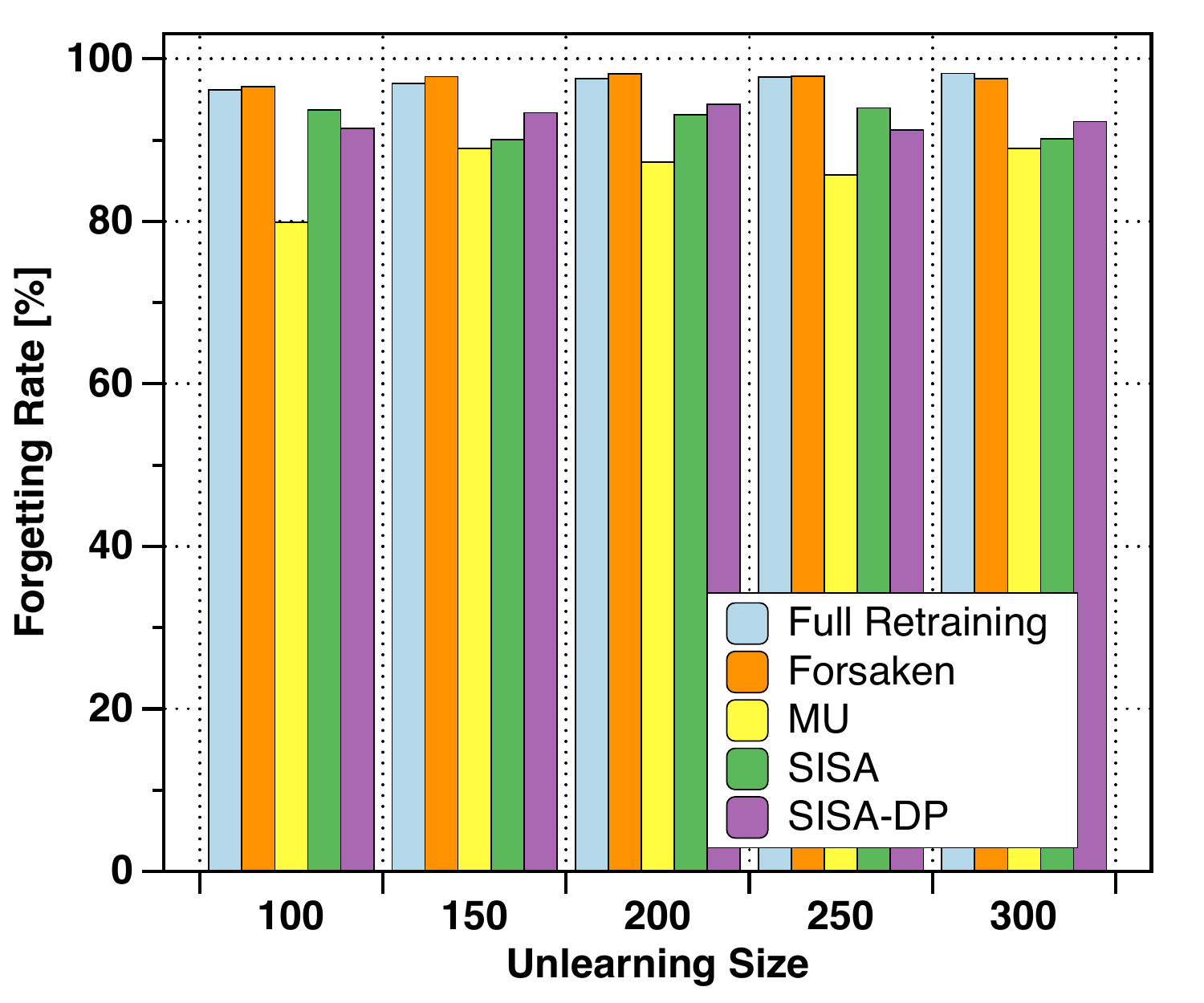}
\label{fig_unlearned_size_cifar_acc}
\end{minipage}
}%
\hfill
\subfigure[The change of Test.Acc with different unlearning sizes for C10.T.]{
\begin{minipage}[t]{0.45\linewidth}
\centering
\includegraphics[scale=0.24]{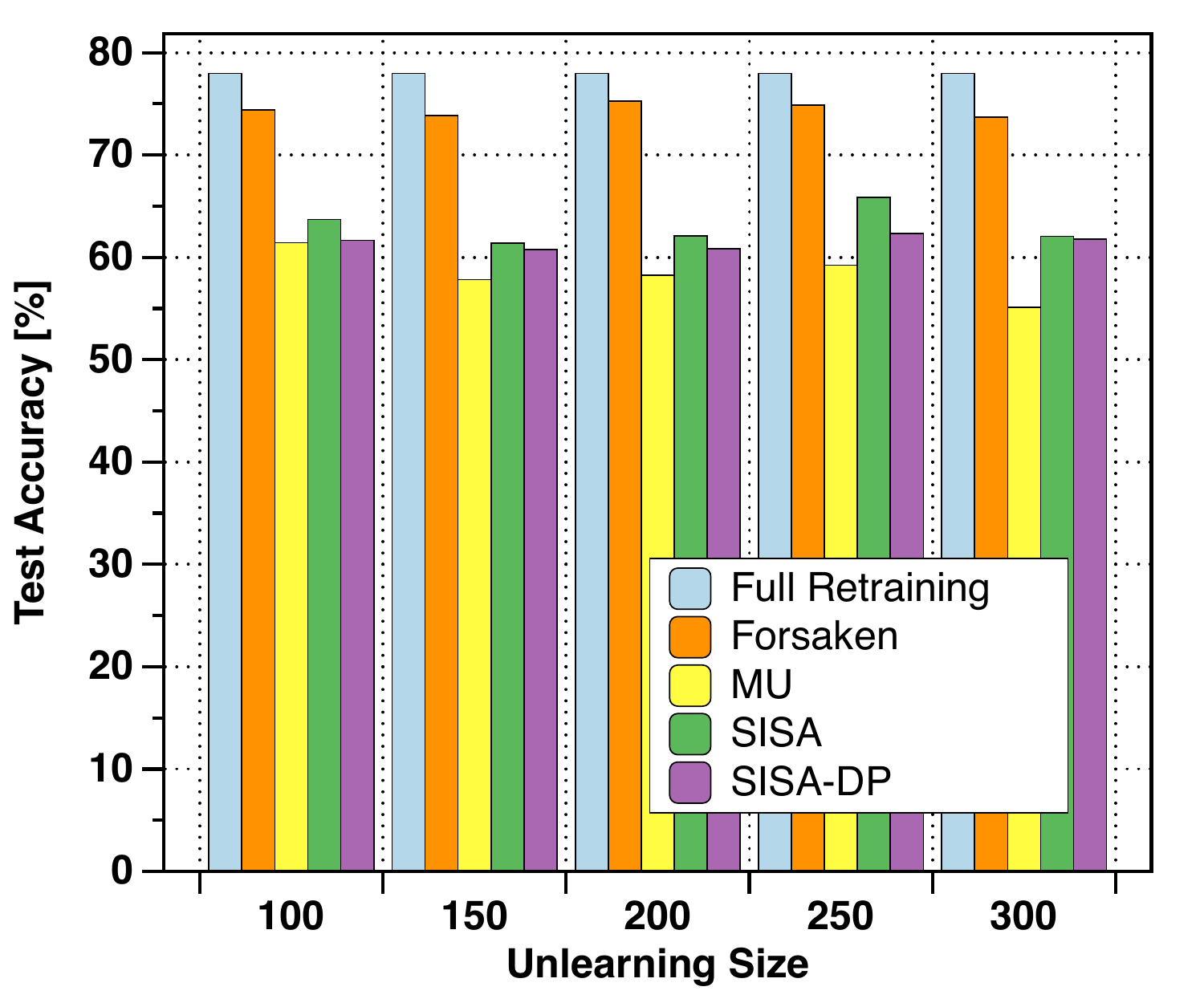}
\label{fig_unlearned_size_cifar_fr}
\end{minipage}
}%
\hfill
\subfigure[The change of $FR$ with different unlearning sizes for I.C.]{
\begin{minipage}[t]{0.45\linewidth}
\centering
\includegraphics[scale=0.24]{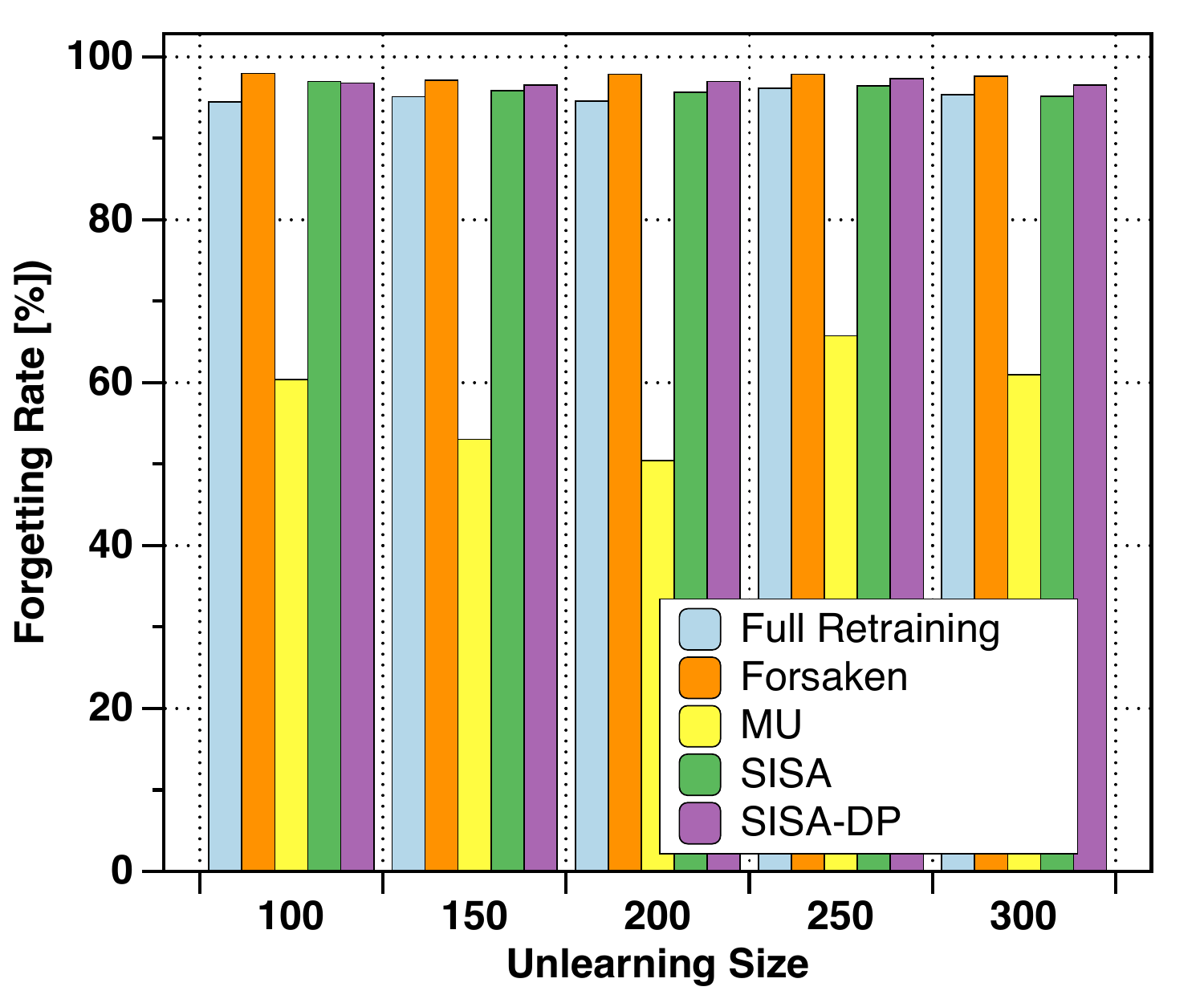}
\label{fig_unlearned_size_imdb_acc}
\end{minipage}
}%
\hfill
\subfigure[The change of Test.Acc with different unlearning sizes for I.C.]{
\begin{minipage}[t]{0.45\linewidth}
\centering
\includegraphics[scale=0.24]{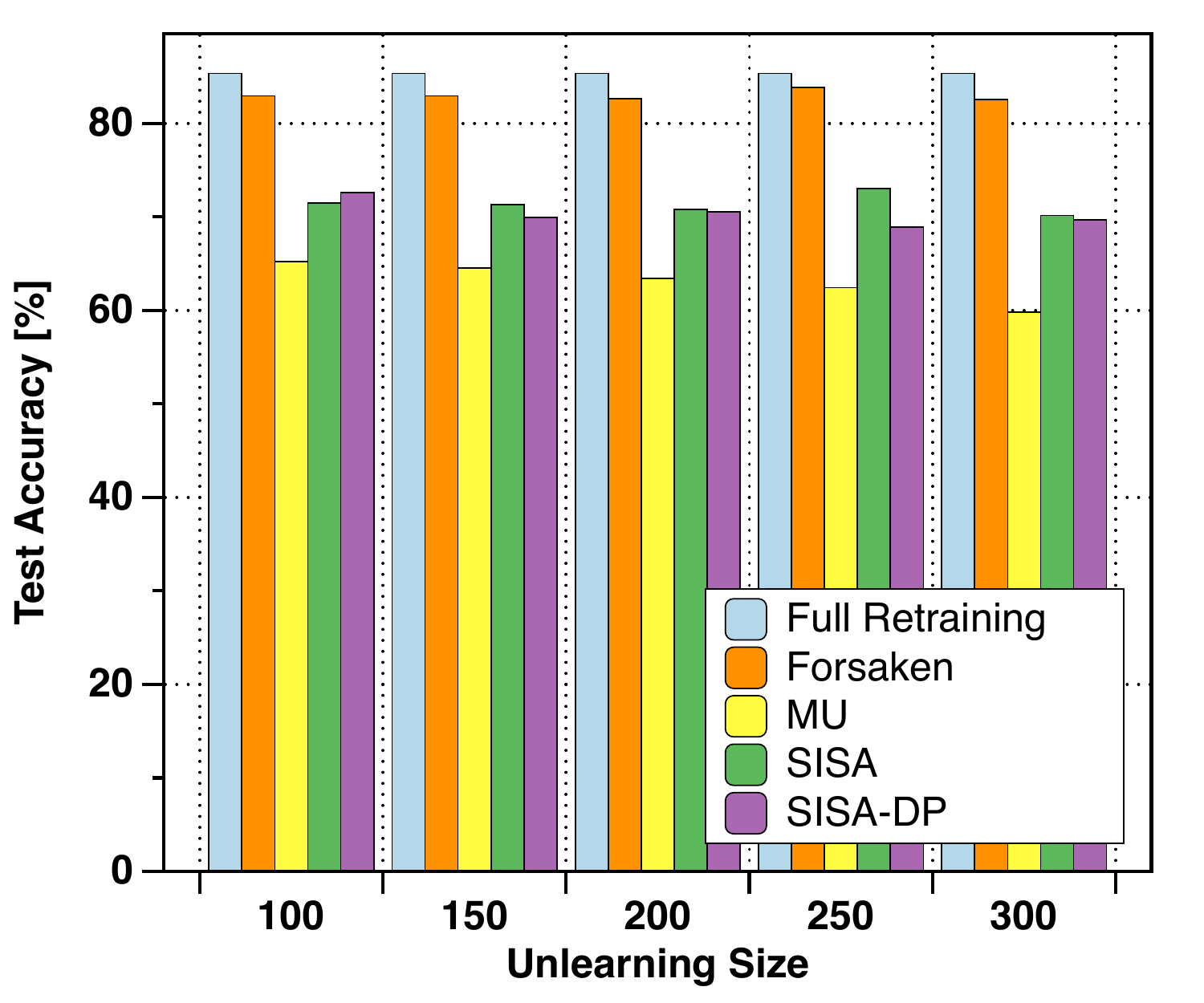}
\label{fig_unlearned_size_imdb_fr}
\end{minipage}
}%
\caption{
\revision{The performance change with different unlearning sizes for OOD data unlearning.
Test.Acc means the testing accuracy. 
}
}
\label{fig_unlearned_size}
\end{figure}

\vspace{0.1cm}
\noindent
\textbf{Efficiency.}
\revision{\textlabel{Except for performance drops, efficiency also counts a lot for applying machine unlearning.}{rB3:1}
Since the computation complexities of OOD and ID data unlearning are identical, we only show the running time comparison result of OOD data unlearning in Table~\ref{table_efficiency_comparison}.
It can be observed that the running time of SISA and SISA-DP is similar to full retraining, because SISA degrades to the performance of retraining the whole model as the percentage of unlearning data is more than 0.075\% (detailedly discussed in~\cite{bourtoule2019machine}), \textlabel{and has to retrain all 10 shards.}{rD3:2}
Compared with the retraining based methods, \sysname can save hundreds of times of overheads.
Even for SMU, which only needs one epoch of training, its running time is several times more than \sysname because the involved training data of \sysname is less than SMU.
To further understand the efficiency comparison results, we theoretically analyze the time complexity as follows.}

Take the sizes of the training set and unlearning set as $N$ and $n_0$, respectively.
Suppose the time required to update the target model with one sample is $\mathcal{O}(1)$.
\revision{The time complexity of full retraining is $\mathcal{O}(K N)$ where $K$ is the number of training epochs.}
The time complexity of \sysname is $\mathcal{O}(t n_0)$, where $t$ is the training iterations of mask gradient generator.
\revision{The time complexity of both SISA and SISA-DP is $\mathcal{O}(K (N - n_0))$.}
SMU only needs one normal epoch of training used to subtract recorded gradients.
Thus, the time complexity of SMU is $\mathcal{O}(N - n_0)$.
Due to $n_0 \ll N$, the running time of \sysname is less than SMU.

\begin{table*}[!htbp]
\centering
\footnotesize
\caption{\revision{Efficiency comparison (average / variance)}}
\resizebox{0.57\textwidth}{!}{%
\begin{tabular}{c|c|c|c|c|c}

\toprule
\multirow{2}{*}{\centering \textbf{Dataset}} & \multicolumn{5}{c}{\textbf{Running Time (s)}} \\


\cline{2-6}
& \textbf{Retraining}
& \textbf{\sysname}
& \centering \textbf{SMU} 
& \textbf{SISA}
& \textbf{SISA-DP} \\


\hline
C10.S.          & 1087.6 / 2.78 & 15.09 / 0.18 & 45.7 / 0.21 & 1000.59 / 1.73 & 1009.61 / 1.63  \\

\hline
C10.T.          & 1089.4 / 2.56 & 16.12 / 0.07 & 47.05 / 0.06 & 1002.25 / 1.85 & 1002.31 / 2.01  \\

\hline
C100.T.         & 925.65 / 1.73 & 7.69 / 0.12 & 61.05 / 0.11 & 851.59 / 2.12 & 852.29 / 1.32 \\

\hline
I.C.            & 372.3 / 0.72 & 2.35 / 0.07 & 10.5 / 0.07 & 342.52 / 0.22 & 345.45 / 1.06 \\

\hline
Reuters (35-11) & 121.6 / 0.67 & 1.68 / 0.02 & 3.35 / 0.03 & 111.87 / 0.14 & 111.12 / 1.32  \\

\hline
News (15-5)     & 174 / 0.73 & 2.57 / 0.09 & 3.91 / 0.05 & 160.08 / 0.26 & 162.72 / 0.49 \\

\bottomrule
\end{tabular}
}
\label{table_efficiency_comparison}
\end{table*}

\vspace{0.1cm}
\noindent
\textbf{Comparison with Different Unlearning Sizes.}
\revision{Overall, \sysname \textlabel{attains the best performance among all baselines.}{rB1:1}
To further evaluate the stability of \sysname, we compare it with other methods with varying unlearning sizes (100 to 300) over two types of datasets, C10.T and IMDB.
Fig.~\ref{fig_unlearned_size} plots the experiment results for OOD data unlearning, and the experiment about ID data unlearning is attached in Appendix~\ref{sec_appendix_unlearning_id}.
The increased unlearning data size only leads to a modest reduction of $FR$ and accuracy for \sysname.
Besides, the performance of \sysname is always close to full retraining and outperforms other methods with different unlearning sizes.
The phenomenon validate that \sysname can perform as stably as retraining-based methods.}

\vspace{0.1cm}
\noindent
\textbf{Elaborate Comparison with SISA.}
\revision{As the retraining-based MU is the most competitive method among all baselines, Table~\ref{table_comparison_SISA} further \textlabel{elaborates comparison of}{rD3:1} \sysname with SISA and SISA-DP.
In more detail, we show the performance change of the three methods with different numbers of shards for OOD data unlearning.
From the experimental results, SISA and SISA-DP attain similar $FR$ to \sysname but cause more accuracy drops as the number of shards increases from $1$ to $4$.
The phenomenon illustrates that although the memorization of OOD data can be forgotten by SISA, the target model loses its utility due to the partition of datasets.
Relatively, \sysname does not suffer from the problem.}

\begin{table}[!htbp]
\centering
\footnotesize
\caption{\revision{Detailed comparison of \sysname with SISA on C10.S. Diff.Acc specifies the testing accuracy difference after conducting machine unlearning.}}

\resizebox{0.45\textwidth}{!}{%
\begin{tabular}{c|c|c|c|c|c|c}

\toprule
\multirow{2}{*}{Shards} & \multicolumn{3}{c|}{FR} & \multicolumn{3}{c}{Diff.Acc} \\

\cline{2-7}
& \sysname & SISA & SISA-DP & \sysname & SISA & SISA-DP \\

\hline
1 & \multirow{6}{*}{97.62\%} & 93.45\% & 94.64\% & \multirow{6}{*}{3.34\%} & 0.0\% & 3.58\% \\

\cline{1-1}\cline{3-4}\cline{6-7}
2 &  & 94.05\% & 95.23\% & & 1.07\% & 4.07\% \\

\cline{1-1}\cline{3-4}\cline{6-7}
4 &  & 92.85\% & 94.05\% & & 3.56\% & 5.56\% \\

\cline{1-1}\cline{3-4}\cline{6-7}
6 &  & 92.26\% & 94.64\% & & 7.47\% & 9.47\% \\

\cline{1-1}\cline{3-4}\cline{6-7}
8 &  & 92.26\% & 95.83\% & & 11.56\% & 13.85\% \\

\cline{1-1}\cline{3-4}\cline{6-7}
10 &  & 93.45\% & 94.05\% & & 14.78\% & 15.97\% \\

\bottomrule
\end{tabular}
}
\label{table_comparison_SISA}
\end{table}

Due to the higher data privacy leakage risk caused by OOD data~\cite{carlini2019secret} than ID data, the following experiments of this section will focus more on OOD data unlearning to further validate the effectiveness and robustness of \sysname.



\subsection{Effect Factors}\label{sub_evaluation_factor}
\revision{In this part, we conduct extensive experiments to analyze the key factors that affect \sysname performance.}
Three different types of datasets are mainly involved, including C10.T., I.C. and Reuters (36-11).

\vspace{0.1cm}
\noindent
\textbf{Training Iteration of Generator.}
We first experiment with the performance change of \sysname under different iteration rounds (shown in Fig.~\ref{fig_iteration}).
Commonly, \sysname reaches its best performance after tens of iterations.
Once the best point is reached, more training iterations hardly affect $FR$ but cause more performance degradation of the target model (mainly reflected in accuracy decreasing).
The reason is that according to our design of mask gradient generator, overtraining of \sysname can cause over-unlearning of unrelated memorization, which lowers the performance of the target model.

Moreover, the efficiency of \sysname is also strongly affected by the number of training iterations.
Table~\ref{table_efficiency} shows the running time of \sysname to accomplish machine unlearning of 200 samples with different iteration numbers.
The results show that the running time linearly increases along with the training iterations.
However, even for the neural network with 14.09M parameters, the mask gradient generation can complete 30 iterations in seconds, which is much faster than other retraining based methods.

\begin{figure}[htbp]
\centering
\subfigure[The performance of $\mathcal{G}$ at each iteration for C10.T.]{
\begin{minipage}[t]{0.45\linewidth}
\centering
\includegraphics[scale=0.24]{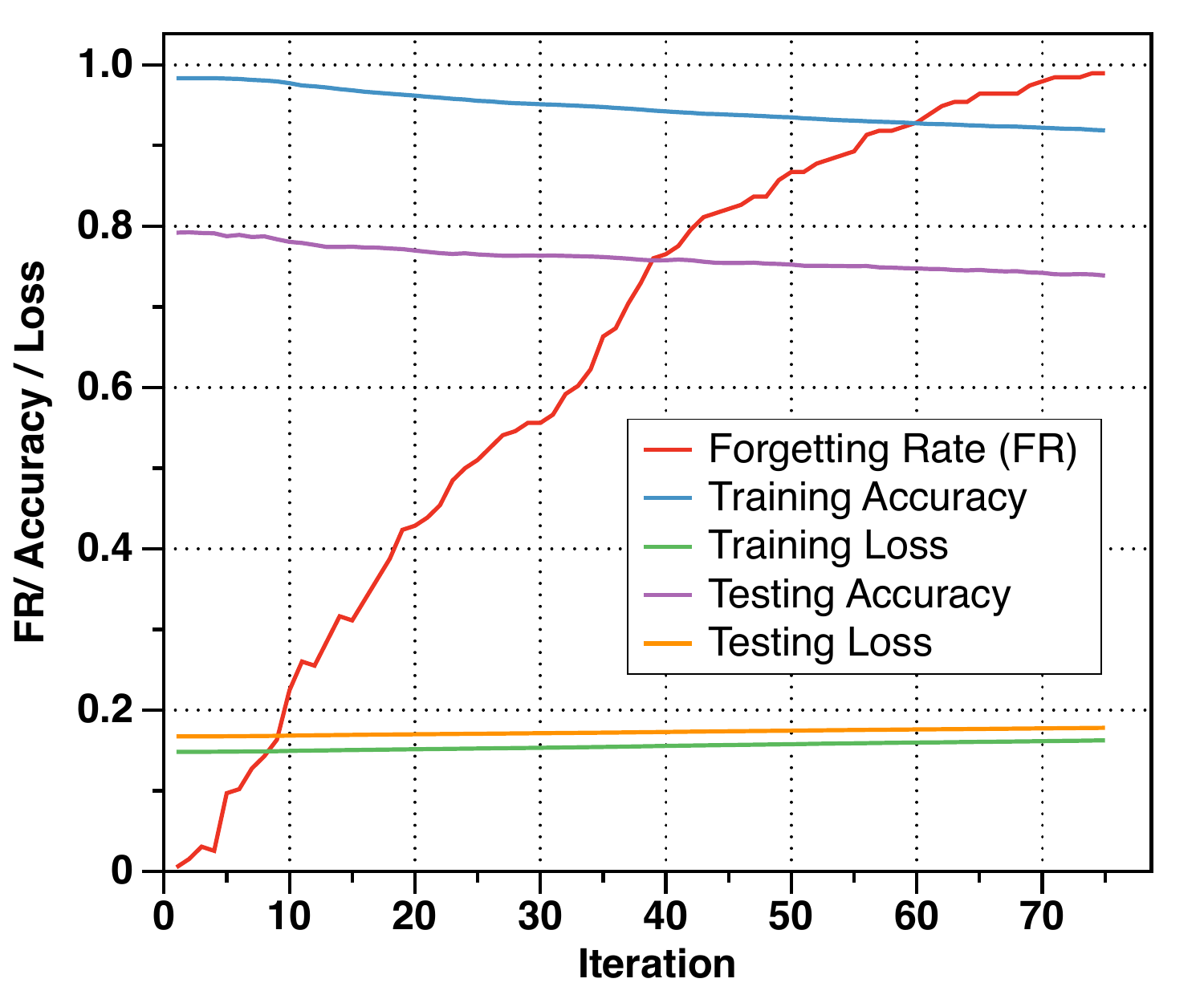}
\label{fig_iteration_cifar}
\end{minipage}
}%
\hfill
\subfigure[The performance of $\mathcal{G}$ at each iteration for I.C.]{
\begin{minipage}[t]{0.45\linewidth}
\centering
\includegraphics[scale=0.24]{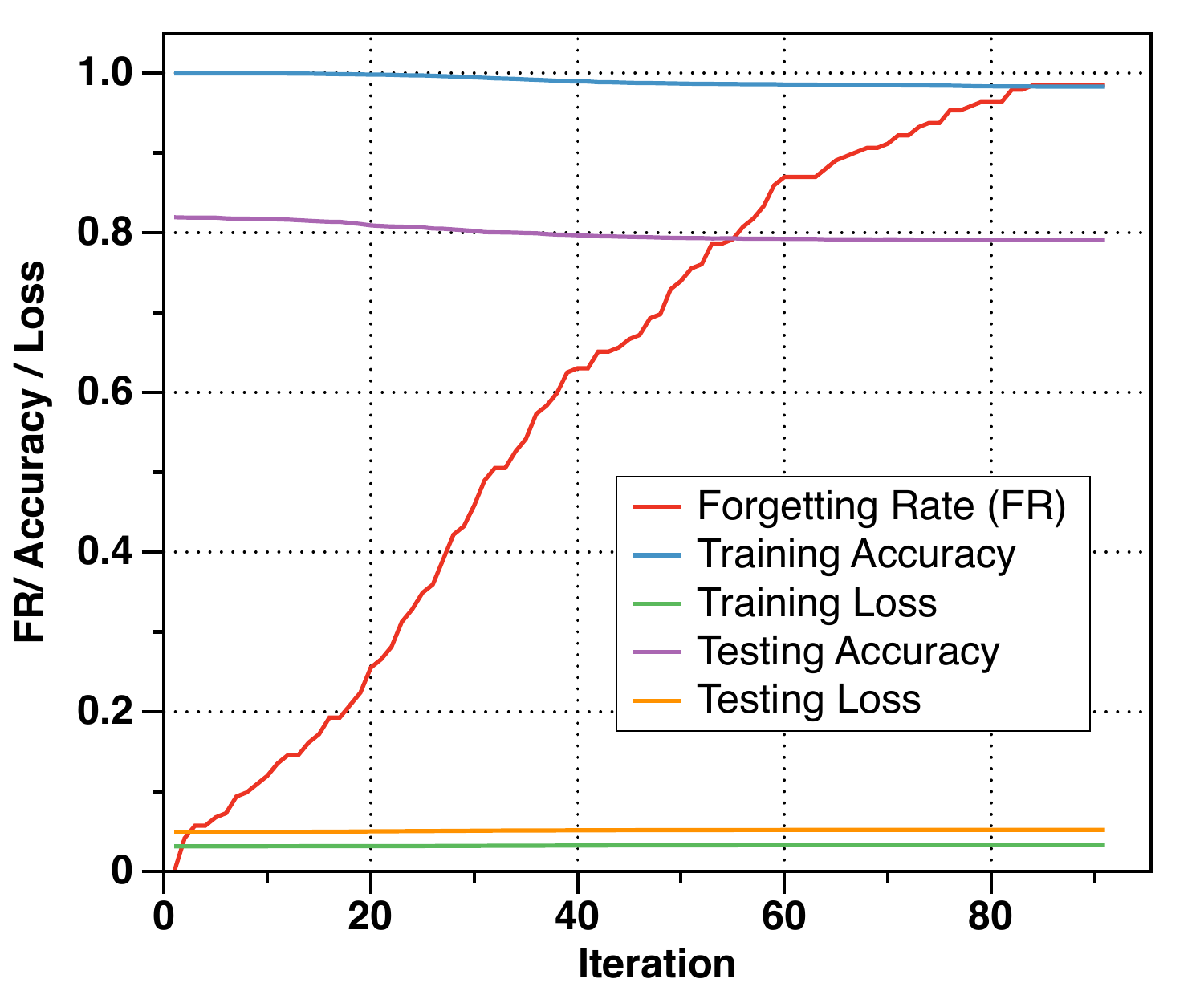}
\label{fig_iteration_imdb}
\end{minipage}
}%
\hfill
\subfigure[The performance of $\mathcal{G}$ at each iteration for Reuters (35-11).]{
\begin{minipage}[t]{0.58\linewidth}
\centering
\includegraphics[scale=0.24]{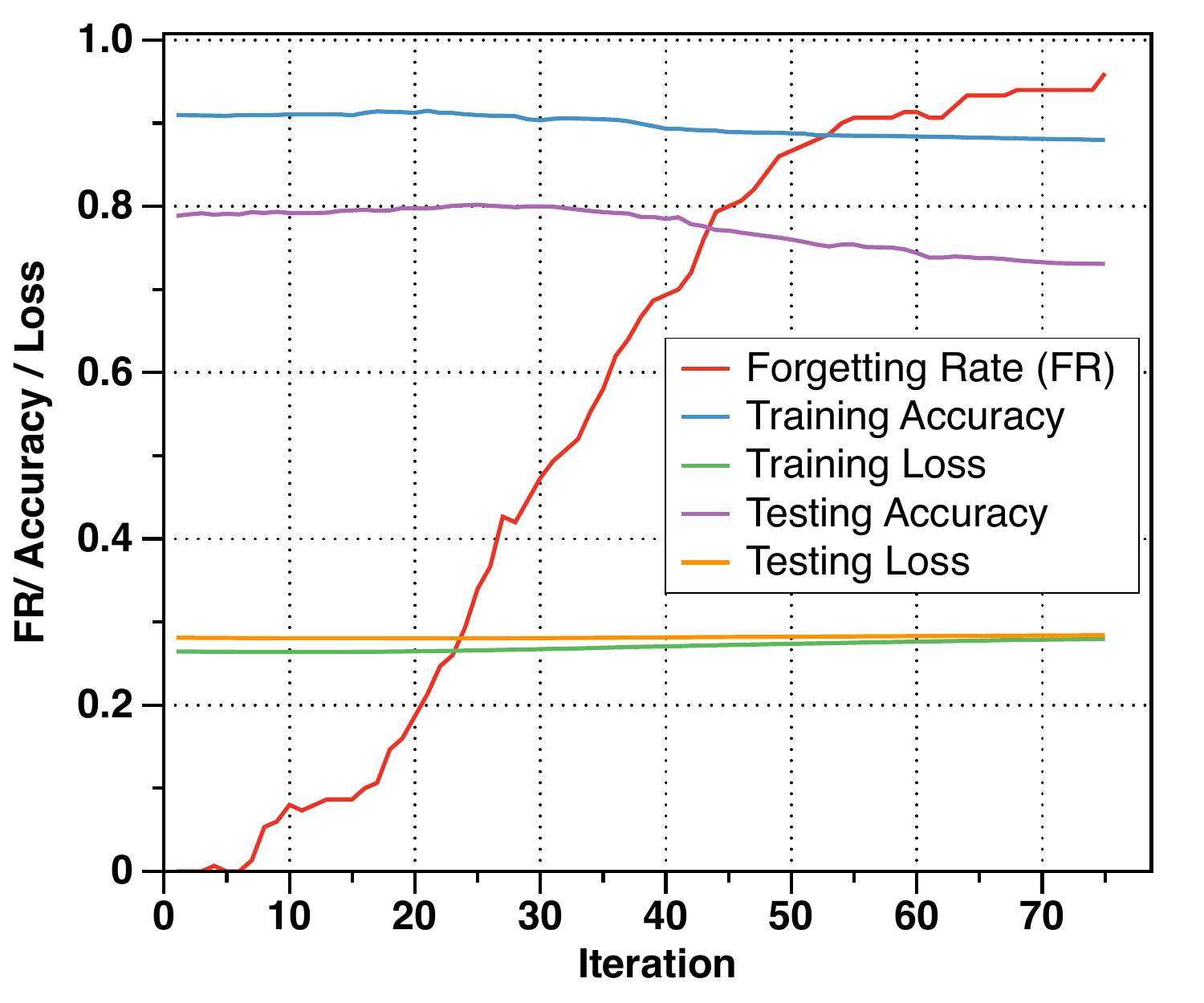}
\label{fig_iteration_reuster}
\end{minipage}
}%
\caption{\revision{The detailed performance evaluation of the mask gradient generator $\mathcal{G}$ at each iteration.}} 
\label{fig_iteration}
\end{figure}

\begin{table}[htbp]
\centering
\footnotesize
\caption{Running time of \sysname with different iterations}
\resizebox{0.33\textwidth}{!}{%
\begin{tabular}{c|c|c|c|c|c}

\toprule
\multirow{2}{*}{\textbf{Iteration}} & \multicolumn{5}{c}{\textbf{Running Time (s)}} \\

\cline{2-6}
& \textbf{20} & \textbf{30} & \textbf{40} & \textbf{50} & \textbf{60} \\

\hline
C10.T.  & 5.21  & 7.44  & 10.31 & 12.46 & 15.93  \\

\hline
I.C.  & 0.82  & 1.28 & 1.68  & 2.02 & 2.47  \\

\hline
R(35-11)  & 0.62  & 0.91 & 1.24 & 1.45 & 1.73 \\

\bottomrule
\end{tabular}
}
\label{table_efficiency}
\end{table}

\vspace{0.1cm}
\noindent
\revision{\textbf{Penalty.}
In \sysname, \textlabel{another important factor is the penalty method used for alleviating catastrophic forgetting.}{rB1:2}
Table~\ref{table_penalty} illustrates the performance of \sysname under different penalty settings.
It can be found that when the penalty is cancelled ($\lambda = 0$), \sysname achieves more than 97\% $FR$ but suffers from about 9\% accuracy drop.
As the weighted penalty mechanism is introduced, accuracy drop is reduced to less than 4\% while maintaining the same level of $FR$.
Moreover, a common phenomenon for the two penalty mechanisms is that both $FR$ and accuracy drop are decreased with the increasing penalty coefficient $\lambda$.
When $\lambda$ is increased to 100, the unlearning process will be blocked, however, the accuracy drop can be well controlled.
Combined with experiments in Section~\ref{sub_exposure}, it is still enough to defend the data leakage caused by unintended memorization even $FR$ is as low as 80\%.
Thus, the choice of high penalty is recommended in applications if there is not a strict restriction on the forgetting rate.}

\begin{table}[!htbp]
\centering
\footnotesize
\caption{\revision{Changing of \sysname performance with different penalty strategies.}}
\resizebox{0.47\textwidth}{!}{%
\begin{tabular}{c|c|c|c|c|c|c}

\toprule
\multicolumn{2}{c|}{\multirow{2}{*}{\textbf{Dataset}}} & \multicolumn{3}{c|}{\textbf{without penalty weight $\omega$}} & \multicolumn{2}{c}{\textbf{with penalty weight $\omega$}} \\

\cline{3-7}
\multicolumn{2}{c|}{}
& \textbf{$\lambda = 0$}
& \textbf{$\lambda = 0.1$}
& \textbf{$\lambda = 1.0$}
& \textbf{$\lambda = 10$}
& \textbf{$\lambda = 100$} \\

\hline
\multirow{2}{*}{C10.T.}
& $FR$
& 97.72\%
& 97.72\%
& 87.75\%
& 98.29\%
& 78.97\% \\

\cline{2-7}
& Diff.Acc
& 8.54\%
& 4.96\%
& 3.5\%
& 3.34\%
& 1.13\% \\

\hline
\multirow{2}{*}{I.C.}
& $FR$
& 98.46\%
& 94.43\%
& 88.72\%
& 98.46\%
& 89.74\% \\

\cline{2-7}
& Diff.Acc
& 6.21\%
& 4.14\%
& 2.32\%
& 2.66\%
& 0.67\% \\

\hline
\multirow{2}{*}{Reuters(35-11)}
& $FR$
& 97.39\%
& 93.39\%
& 85.54\%
& 96.35\%
& 86.98\% \\

\cline{2-7}
& Diff.Acc
& 7.85\%
& 5.59\%
& 2.64\%        
& 4.61\%
& 1.19\% \\

\bottomrule

\multicolumn{7}{c}{Here, Diff.Acc specifies the testing accuracy drops.}\\

\end{tabular}
}
\label{table_penalty}
\end{table}

\begin{figure}[ht!]
\centering
\includegraphics[scale=0.35]{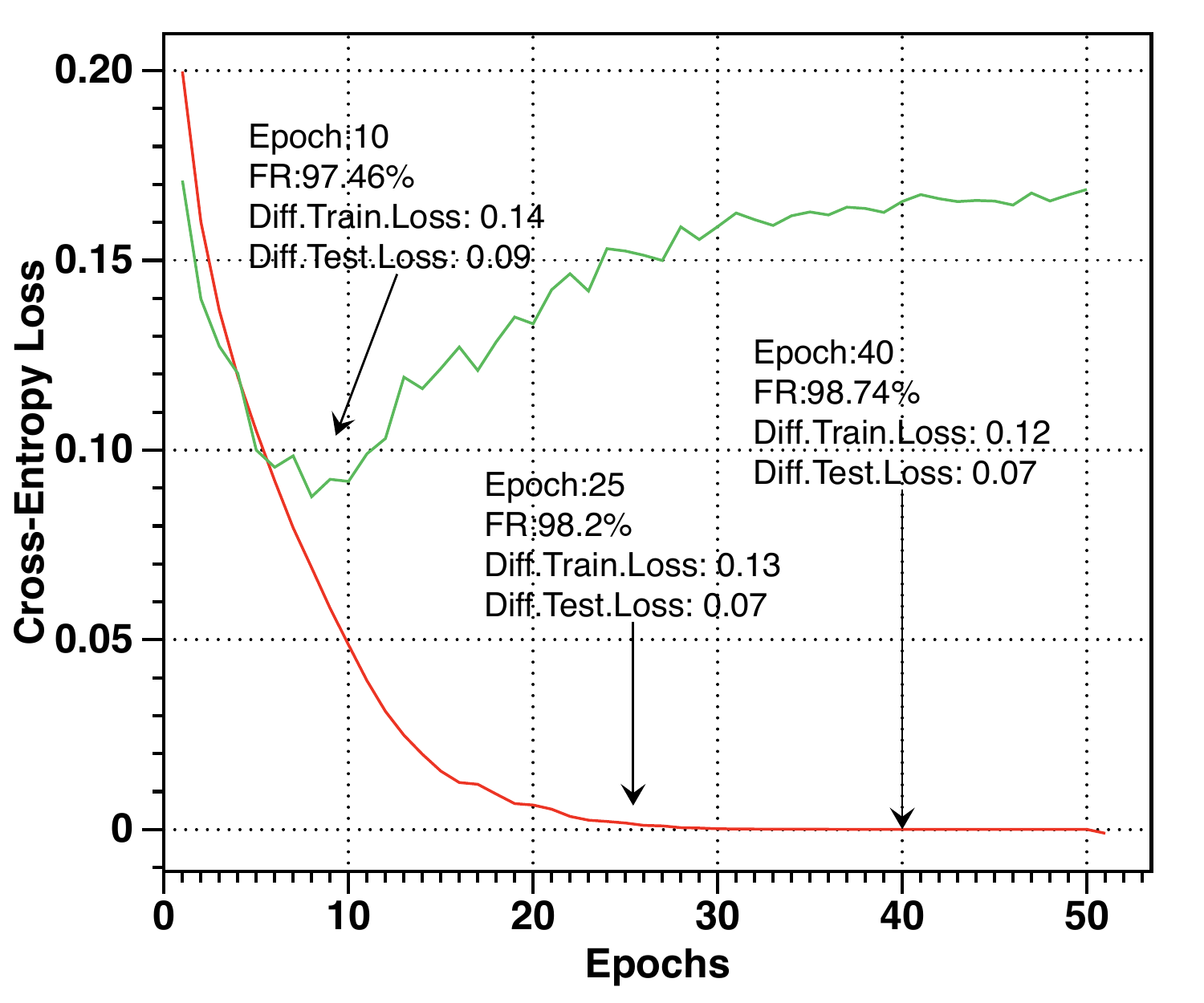}
\caption{The performance of \sysname versus overtraining for C10.T. \sysname is performed at different epochs of training.}
\label{fig_overtraining}
\end{figure}

\noindent
\textbf{Overtraining.}
According to~\cite{arrieta2020explainable, carlini2019secret, salem2018ml}, overtraining is also a factor that tightly relates to machine learning memorization.
Loosely speaking, overtraining brings deeper memorization of the trained model towards the training set.
Fig.~\ref{fig_overtraining} shows a typical example of overtraining with C10.T.
At the first few epochs, the testing loss drops rapidly until reaching the best point.
After the point, the trend of testing loss is reversed, which means the model begins to be overtrained.
To evaluate \sysname versus overtraining, we record the machine unlearning results at different stages of model training, including before and after overtraining.
Every time we measure the performance of \sysname in the experiment, a new membership oracle is trained (training shadow model with the same epochs as the target model).
This is because, unlike the previous experiments, the target model successively changes in the overtraining experiment.
It can be discovered that \sysname performs a little worse when the degree of overtraining is weaker.
The reason is that the deeper memorization of the overtrained model leaves more space for \sysname to fine-tune the target model for machine unlearning.


\begin{figure}[htbp!]
\centering
\subfigure[Visualization of training samples, unknown samples and unlearned samples before machine unlearning for C10.T.]{
\begin{minipage}[t]{0.45\linewidth}
\centering
\includegraphics[scale=0.24]{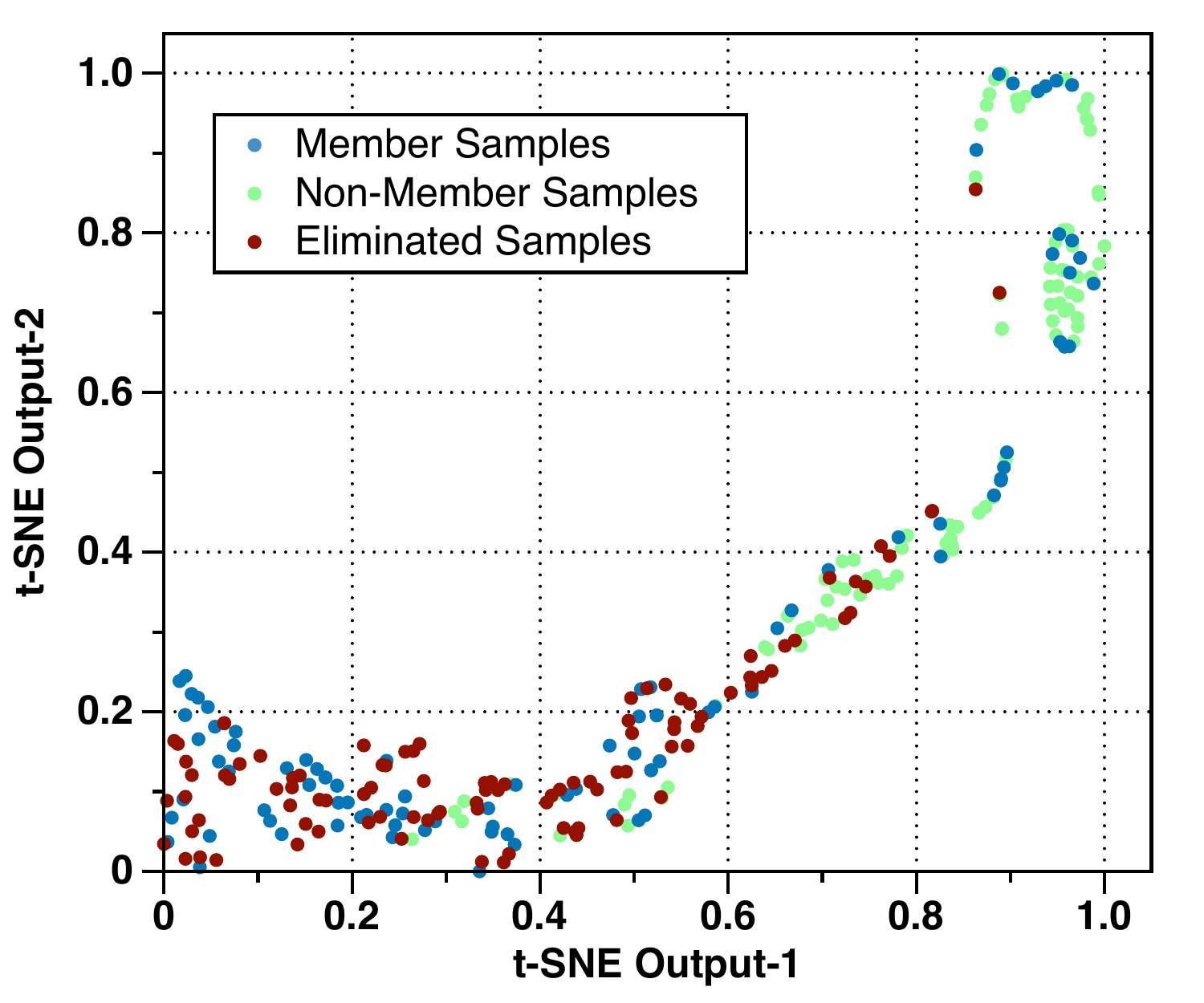}
\label{fig_visualization_cifar10_before}
\end{minipage}
}%
\hfill
\subfigure[Visualization of training samples, unknown samples and unlearned samples after machine unlearning  for C10.T.]{
\begin{minipage}[t]{0.45\linewidth}
\centering
\includegraphics[scale=0.24]{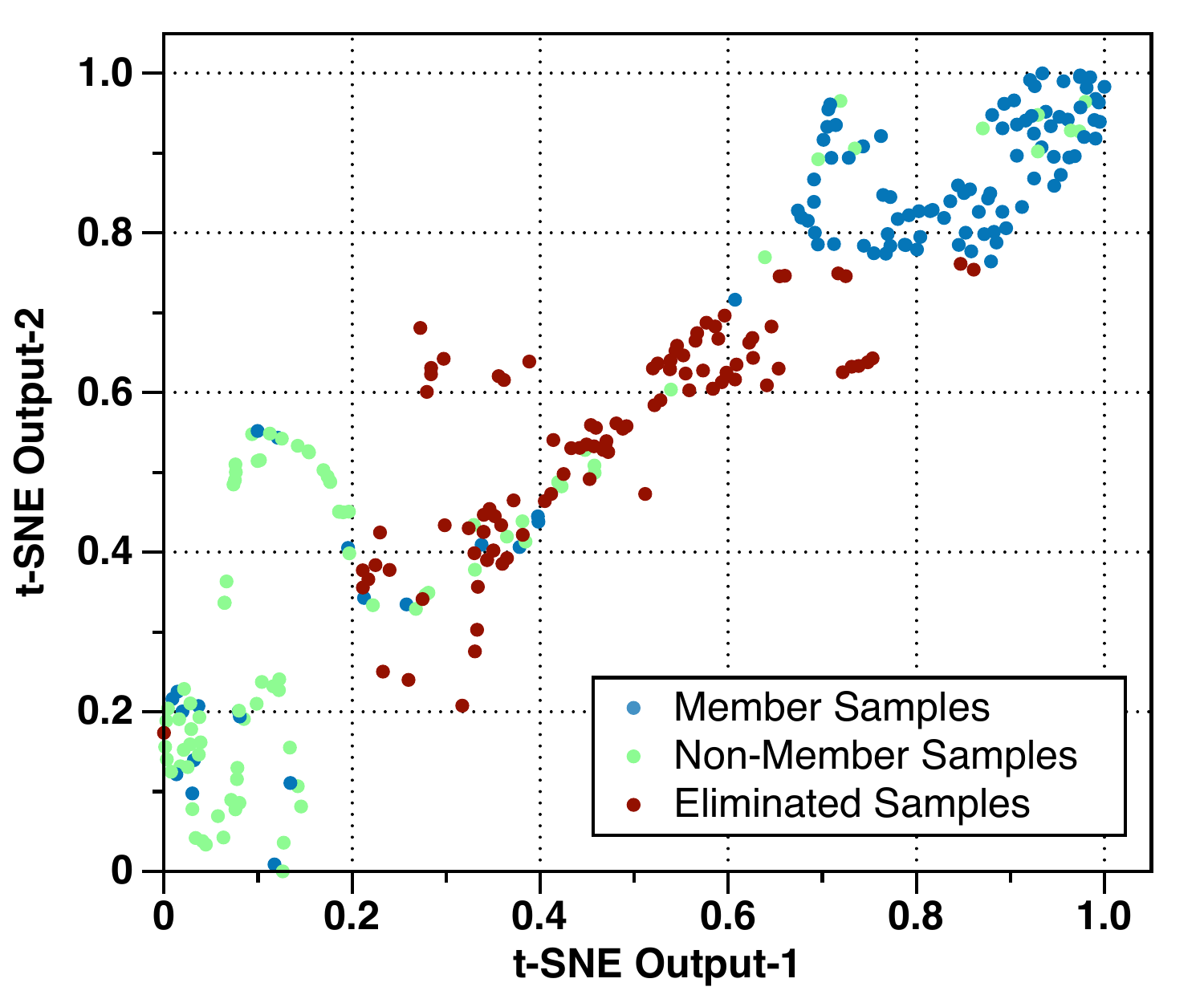}
\label{fig_visualization_cifar10_after}
\end{minipage}
}%
\caption{Visualization of the data attribution before and after conducting machine unlearning for C10.T. In the figures, the location changes of the training and testing samples are mainly caused by the randomization of t-SNE. The labels of the points in the graph are based on the actual values, not membership oracle, and do not change with machine unlearning.}
\label{fig_visualization_cifar10}
\end{figure}

\begin{figure}[htbp]
\centering
\subfigure[Visualization of I.C. before machine unlearning.]{
\begin{minipage}[t]{0.45\linewidth}
\centering
\includegraphics[scale=0.24]{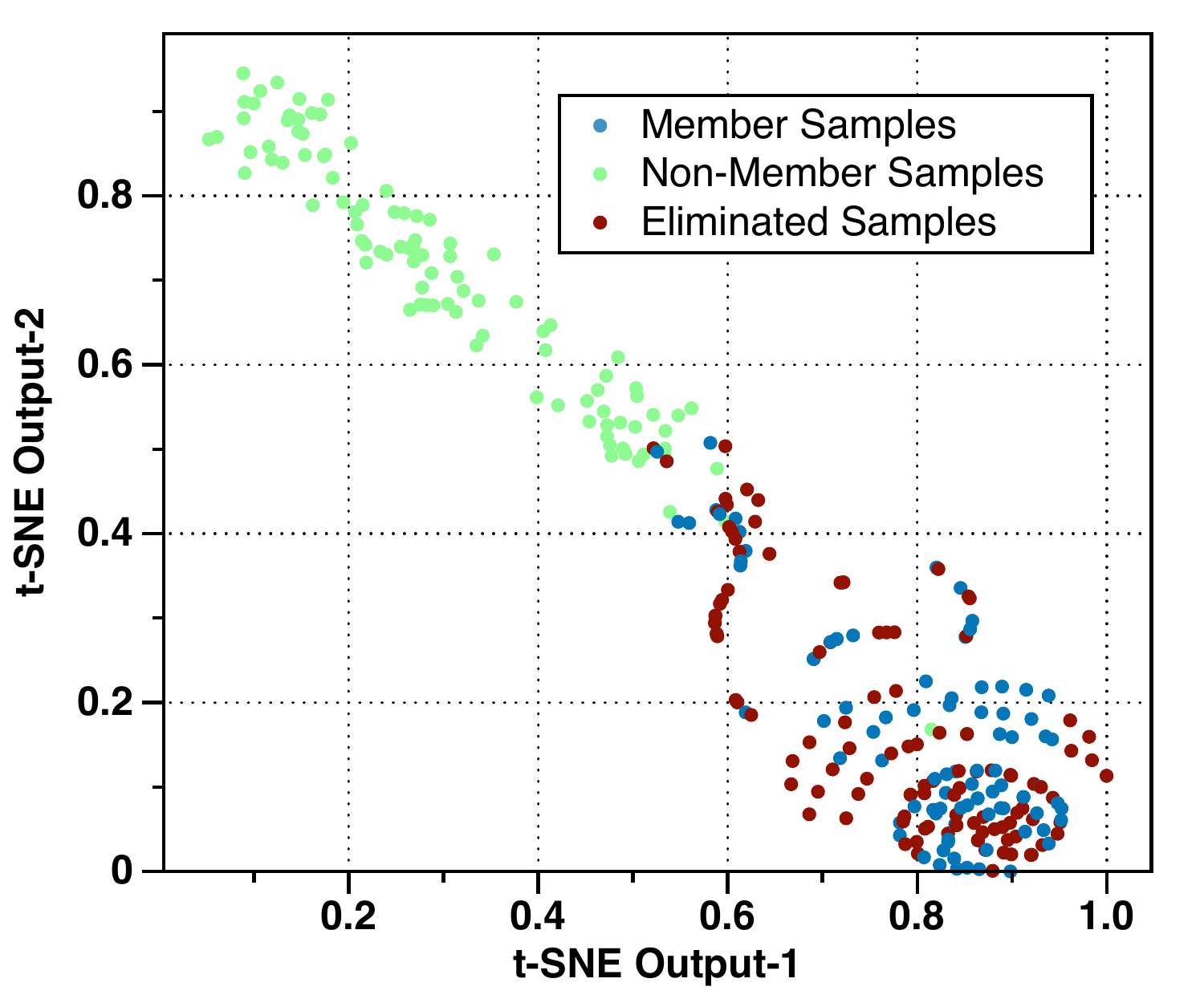}
\label{fig_visualization_imdb_before}
\end{minipage}
}%
\hfill
\subfigure[Visualization of I.C. after machine unlearning.]{
\begin{minipage}[t]{0.45\linewidth}
\centering
\includegraphics[scale=0.24]{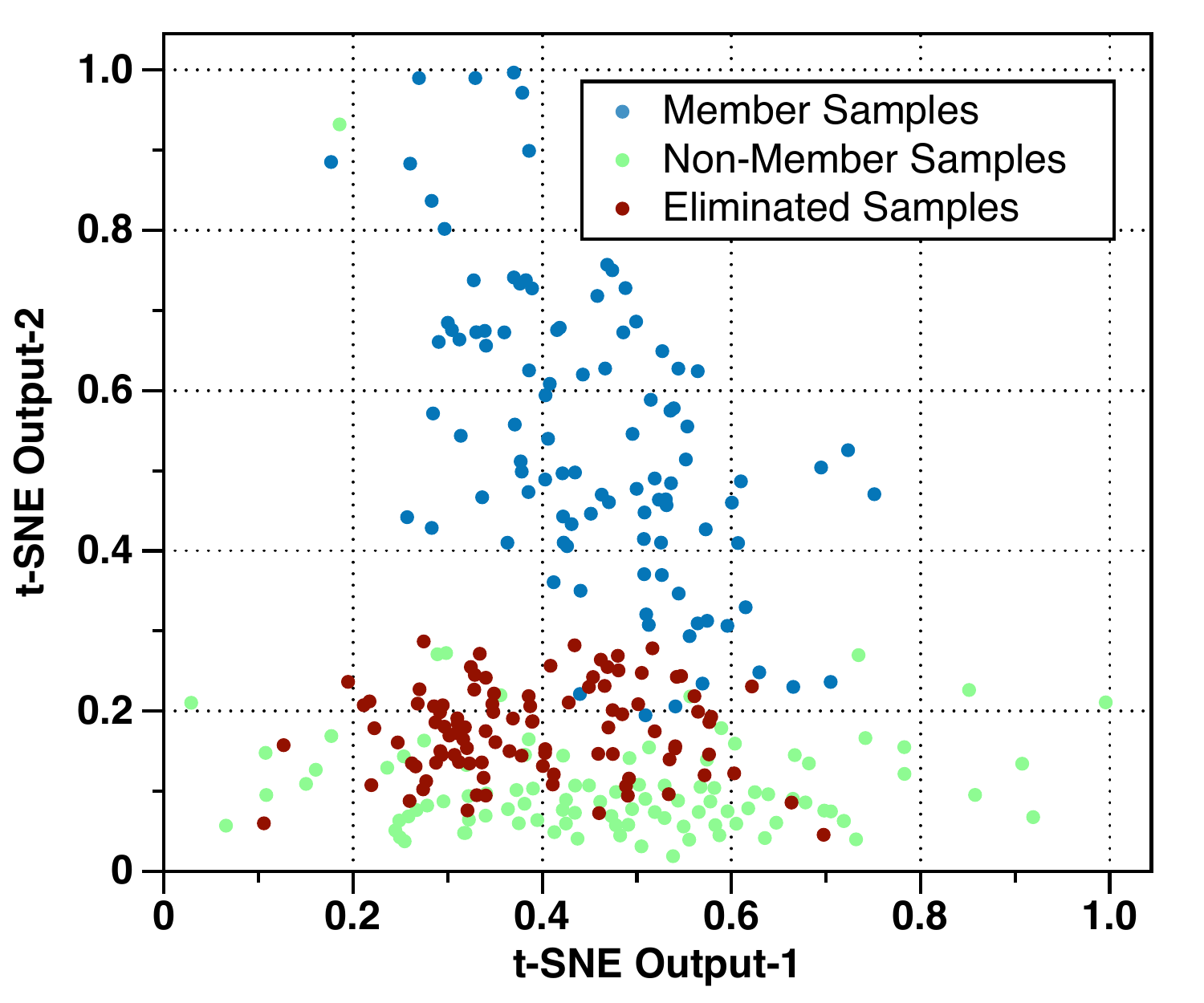}
\label{fig_visualization_imdb_after}
\end{minipage}
}%
\caption{Visualization of the data attribution change before and after conducting machine unlearning for I.C.}
\label{fig_visualization_imdb}
\end{figure}

\subsection{Visualization}
To intuitively evaluate the effectiveness of \sysname, we visualize the change of data attribution before and after performing machine unlearning.
The visualization is based on t-distributed Stochastic Neighbor Embedding(t-SNE), a feature dimension reduction method~\cite{maaten2008visualizing}.
Here, we only focus on the analysis of the OOD data unlearning scenario because \sysname achieves similar performance on both OOD and ID data unlearning.
Two types of learning tasks are picked to conduct the visualization experiment, namely C10.T. and I.C.
Fig.~\ref{fig_visualization_cifar10} and Fig.~\ref{fig_visualization_imdb} demonstrate the visualization results.

Three kinds of data are involved in the experiments, namely 100 member samples from the training set, 100 non-member samples from the testing set and 100 unlearning samples from the OOD data set.
Using t-SNE, we reflect the top three posteriors of all samples into 2-dimension data points~\cite{salem2018ml}.
In the graphs, the colors of these samples are based on their attributions, which means a sample from the testing set is marked as green point, the training sample is the blue point and the eliminated sample is the brown point.
As the dimension embedding of t-SNE is related to the random state and input values, the locations of the training and testing samples before and after machine unlearning are not fixed.
From the results, most of the eliminated samples are close to the group of training samples before machine unlearning.
After machine unlearning, the eliminated samples tend to move to the non-member testing group.
In other words, \sysname successfully makes the eliminated samples change from ``memorized'' to ``unknown''.
Such a phenomenon is in accordance with Definition~\ref{def_k_elimination}, which proves the effectiveness of our methodology to achieve machine unlearning.

\subsection{\revision{White-box Membership Oracle}}
\label{sec_appendix_white_box}
\revision{
    \textlabel{Considering the hardness}{rD2:1} to get access to the whole training set in many application scenarios (e.g., federated learning), the above all experiments are conducted with shadow model based black-box membership oracle (BMI).
    To evaluate our method more comprehensively, we also experiment with the white-box membership oracle (WMI) as the evaluation tool.
    In these experiments, WMI oracles are trained according to the method proposed in~\cite{nasr2019comprehensive}.}

\begin{table}[!htbp]
\centering
\footnotesize
\caption{\revision{Comparison with both black-box and white-box membership oracles.}}

\resizebox{0.36\textwidth}{!}{%
\begin{tabular}{c|c|c}

\toprule
\multirow{2}{*}{\textbf{Dataset}} & \multicolumn{2}{c}{\textbf{Forgetting rate ($FR$)}}\\

\cline{2-3}
         & \textbf{Black-box Oracle} & \textbf{White-box Oracle}\\

\hline
C10.S.   & 97.62\% & 97.02\%  \\

\hline
C10.T.   & 98.29\% & 98.29\% \\

\hline
C100.T.  & 95.73\% & 96.58\%  \\

\hline
I.C.     & 98.46\% & 97.44\%  \\

\hline
Reuters (35-11) & 96.35\% & 96.88\%  \\

\hline
News (15-5)     & 97.53\% & 96.29\%  \\

\bottomrule
\end{tabular}
}
\label{table_white_box}
\end{table}
    
\revision{
    From Table~\ref{table_white_box}, it can be found that with WMI oracles, \sysname achieves close performance to the results computed with BMI.
    To further validate such a statement, we further conduct visualization analysis with WMI oracles, the result of which is given in Fig.~\ref{fig_visualization_white_box_cifar} and Fig.~\ref{fig_visualization_white_box_imdb}.
    For WMI based visualization, we change the encoder (embedding) layer output size in~\cite{nasr2019comprehensive} from $1$ to be $10$ and apply the sigmoid function to map the embedding outputs to range $[0, 1]$.
    With t-SNE, the top-3 embedding outputs are further mapped to $2$-dimension points for visualization.
    From the result, it can be observed that similar to BMI, most eliminated data points are mixed together with other member data points before machine unlearning.
    Then, after \sysname is conducted, the eliminated data points are moved from the member area to the non-member area.
    The fact proves that our proposed indicator and \sysname are feasible and robust with no matter BMI or WMI as the evaluation tool.
}


\begin{figure}[htbp]
\centering
\subfigure[Visualization of training samples, unknown samples and unlearned samples before machine unlearning for C10.T.]{
\begin{minipage}[t]{0.45\linewidth}
\centering
\includegraphics[scale=0.24]{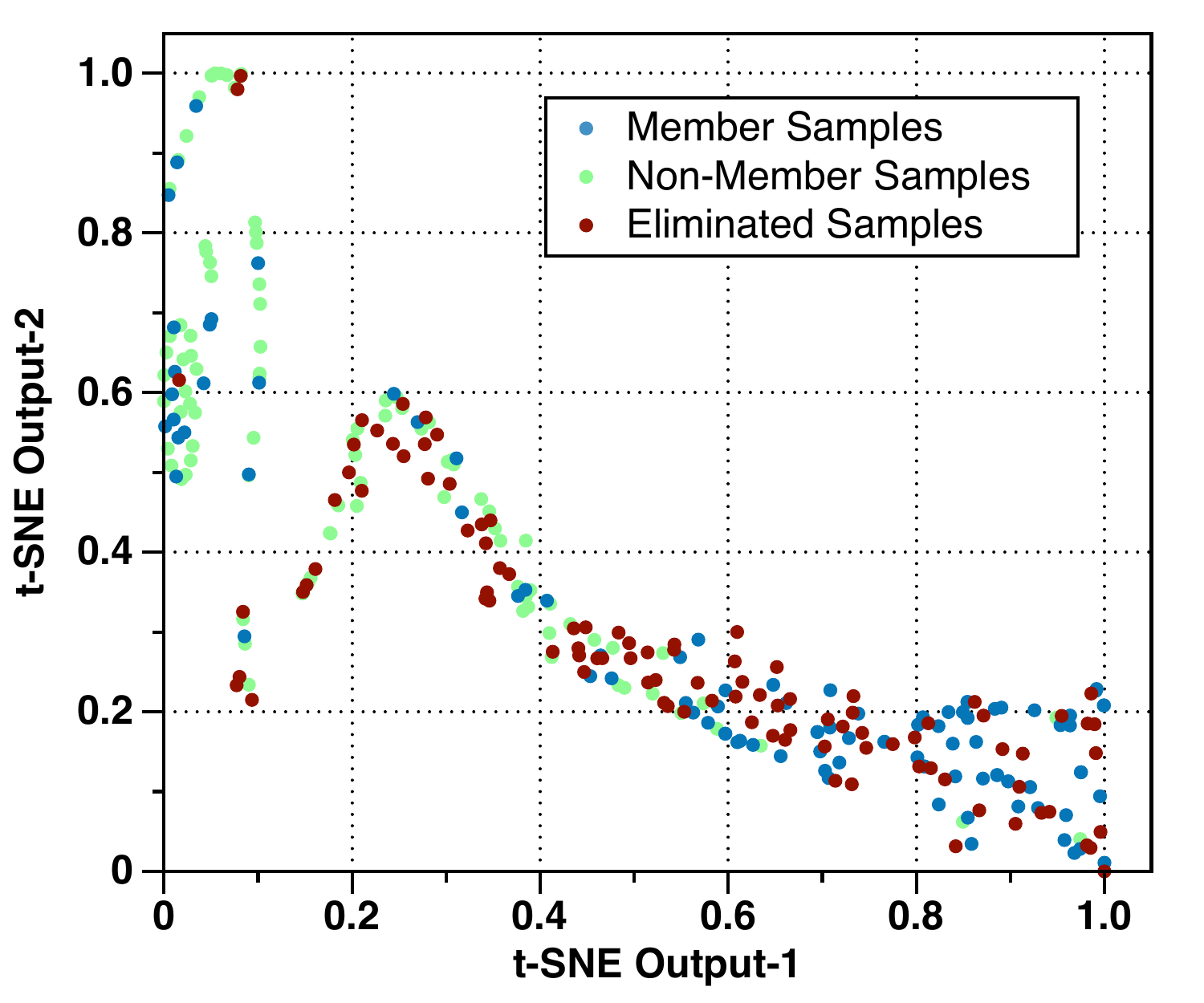}
\label{fig_visualization_white_box_cifar_before}
\end{minipage}
}%
\hfill
\subfigure[Visualization of training samples, unknown samples and unlearned samples after machine unlearning for C10.T.]{
\begin{minipage}[t]{0.45\linewidth}
\centering
\includegraphics[scale=0.24]{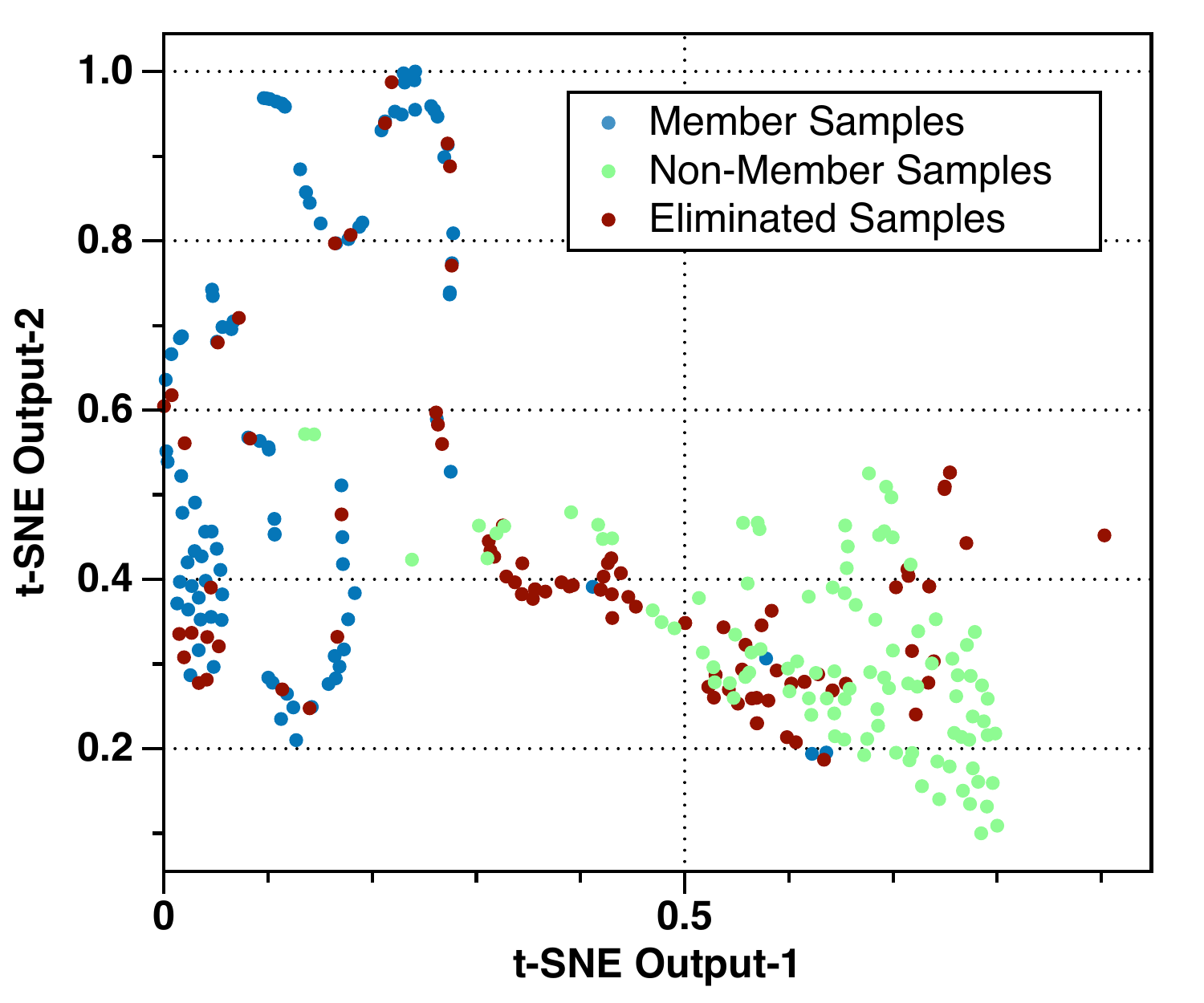}
\label{fig_visualization_white_box_cifar_after}
\end{minipage}
}%
\caption{\revision{Visualization of the data attribution before and after machine unlearning for C10.T with white-box membership oracle.}}
\label{fig_visualization_white_box_cifar}
\end{figure}

\begin{figure}[htbp]
\centering
\subfigure[Visualization of training samples, unknown samples and unlearned samples before machine unlearning for I.C.]{
\begin{minipage}[t]{0.45\linewidth}
\centering
\includegraphics[scale=0.24]{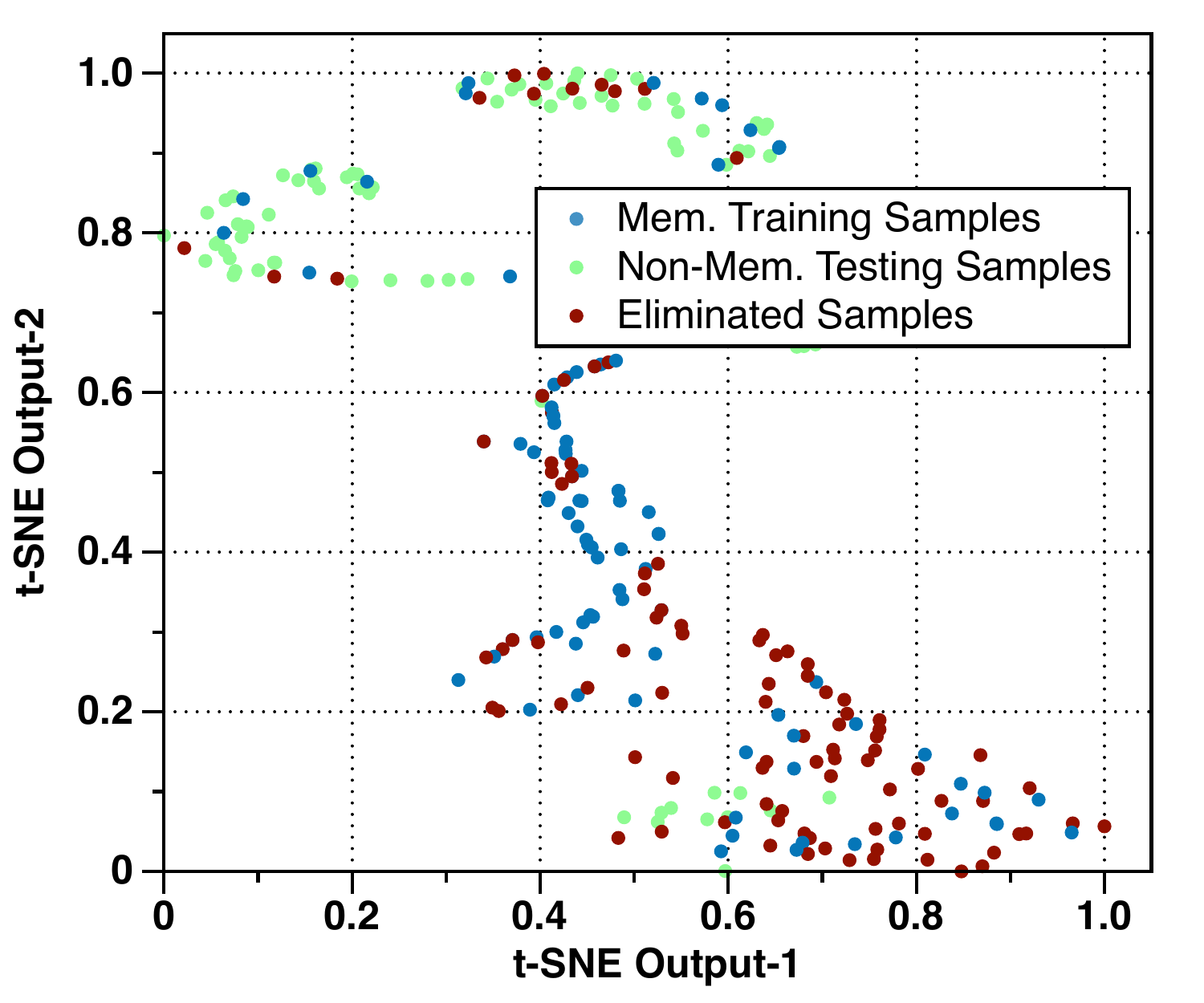}
\label{fig_visualization_white_box_imdb_before}
\end{minipage}
}%
\hfill
\subfigure[Visualization of training samples, unknown samples and unlearned samples after machine unlearning for I.C.]{
\begin{minipage}[t]{0.45\linewidth}
\centering
\includegraphics[scale=0.24]{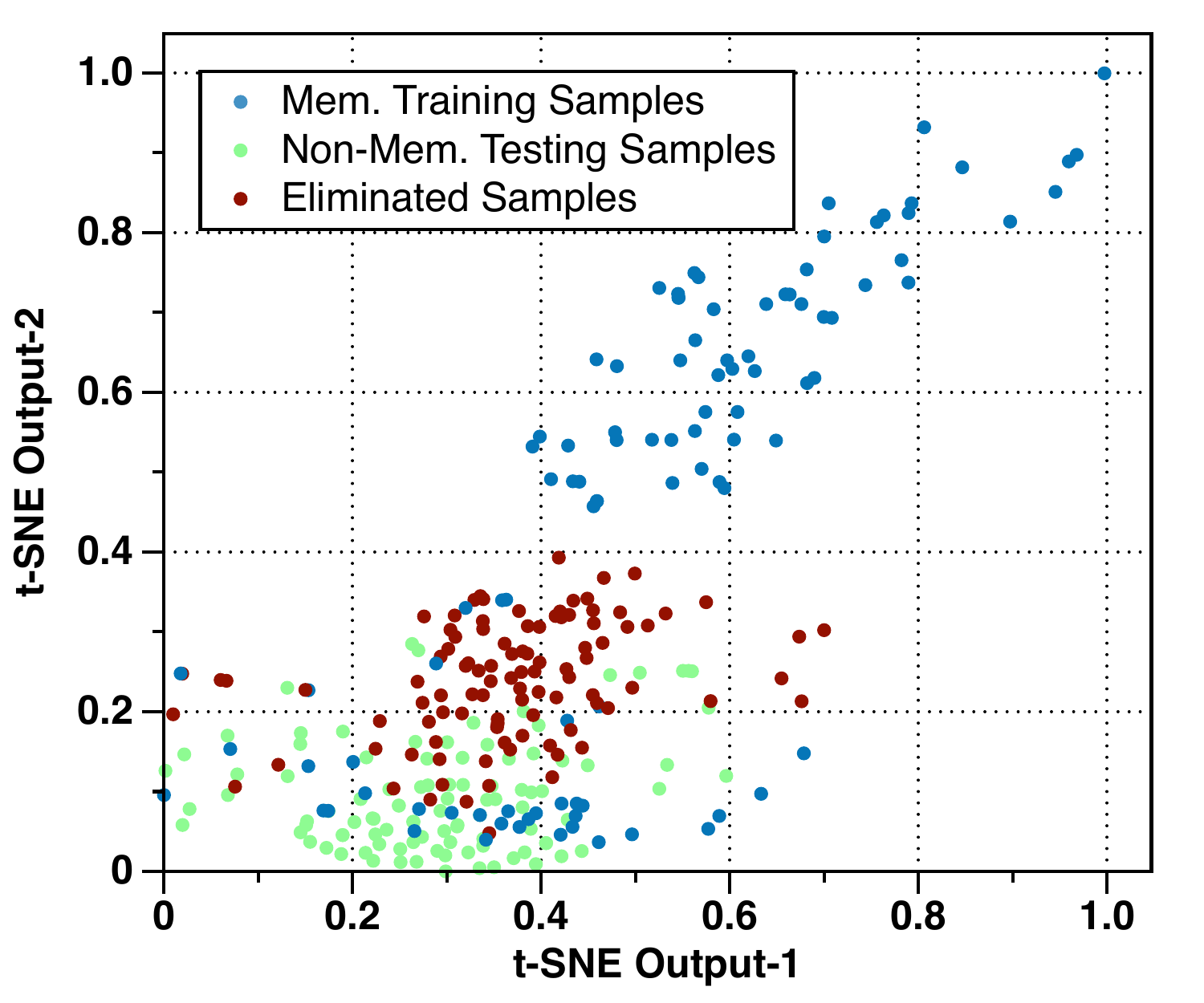}
\label{fig_visualization_white_box_imdb_after}
\end{minipage}
}%
\caption{\revision{Visualization of the data attribution before and after machine unlearning for I.C with white-box membership oracle.}}
\label{fig_visualization_white_box_imdb}
\end{figure}

\subsection{\revision{Poison Data}}
\label{sec_appendix_poison_data}
\revision{\textlabel{To further validate the robustness of}{rC1:1} \mbox{\sysname}, we expand \mbox{\sysname} to the poison data forgetting scenario proposed in~\mbox{\cite{tolpegin2020data}}.
Here, the goal of machine unlearning is to improve the target model's performance by eliminating polluted memories.
In the experiments, the target dataset is CIFAR10.
The percentage of introduced poison data is 2\%, 4\%, 10\%.
The attack adopts the $5\to3$ setting in~\mbox{\cite{tolpegin2020data}}, which means parts of data that should be labeled as $5$ are modified to be $3$ during the model training process.
Moreover, for OOD or ID data unlearning, non-member sample distribution $\mathcal{P}$ is derived from non-member sample posteriors.
For poison data unlearning, $\mathcal{P}$ is formed by the posteriors of the original data without poisoned labels.
In other words, the learning task of \sysname is changed from eliminating useless memories to correcting polluted memories.
From Fig.~\mbox{\ref{fig_poison_data}}, it can be observed that the introduction of poison data can cause about 1\% accuracy drop and 8.4\% recall rate drop compared to the original model (full retraining).
SMU, SISA and SISA-DP fail to achieve performance improvement because of their native drawbacks as mentioned before.
\sysname fails to improve the accuracy of the target model but successfully improves the recall rate.
Since recall rate quantifies the number of correct positive predictions made out of all positive predictions, \sysname indeed improves the performance of the target model over the polluted category.
From the above, we justify the generality of \sysname to process polluted data.}

\begin{figure}[htbp]
\centering
\subfigure[\revision{Test accuracy change for poison data forgetting.}]{
    \begin{minipage}[t]{0.45\linewidth}
    \centering
    \includegraphics[scale=0.24]{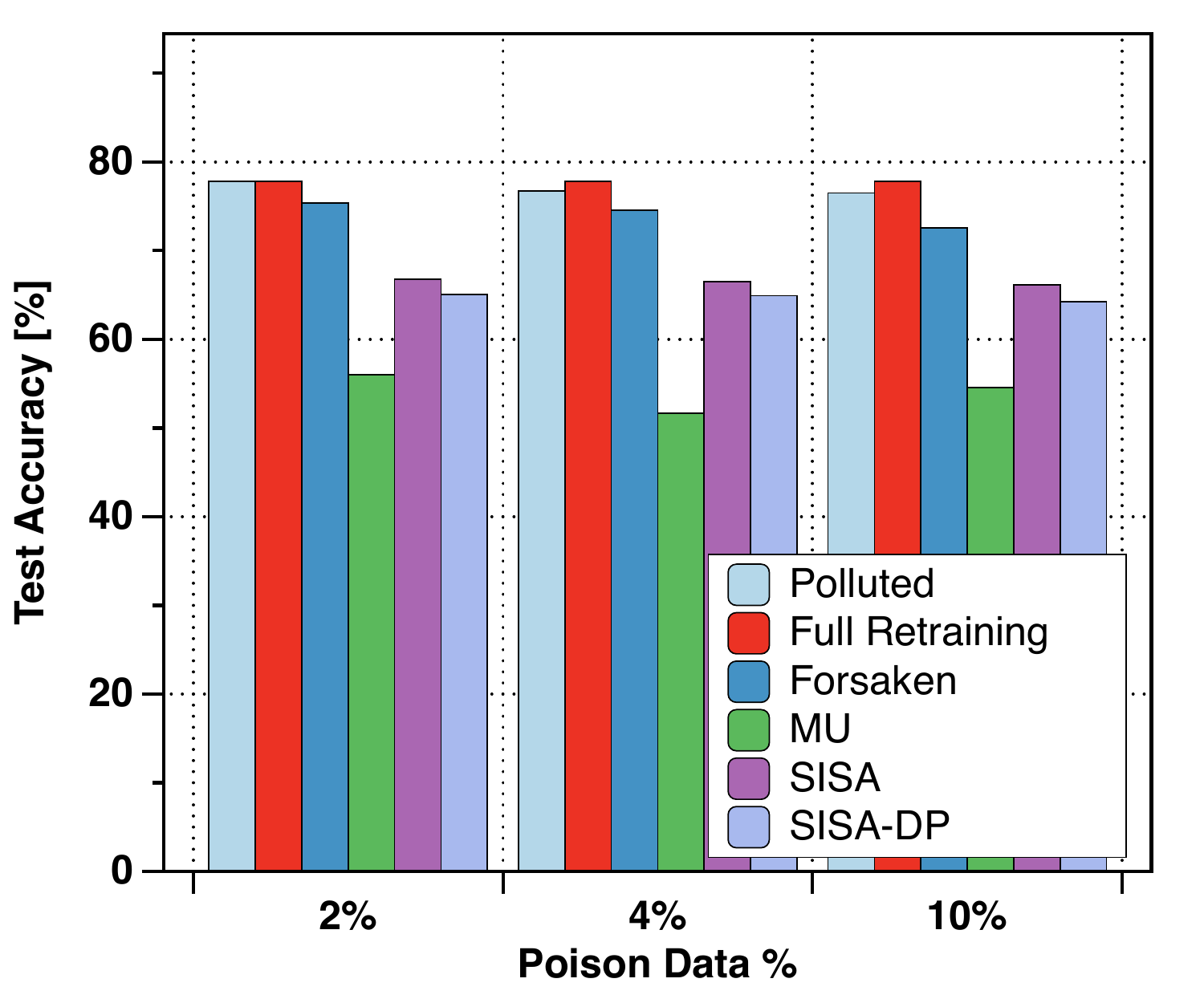}
    \label{fig_poison_data_acc}
    \end{minipage}
}%
\hfill
\subfigure[\revision{Recall rate change for poison data forgetting.}]{
    \begin{minipage}[t]{0.45\linewidth}
    \centering
    \includegraphics[scale=0.24]{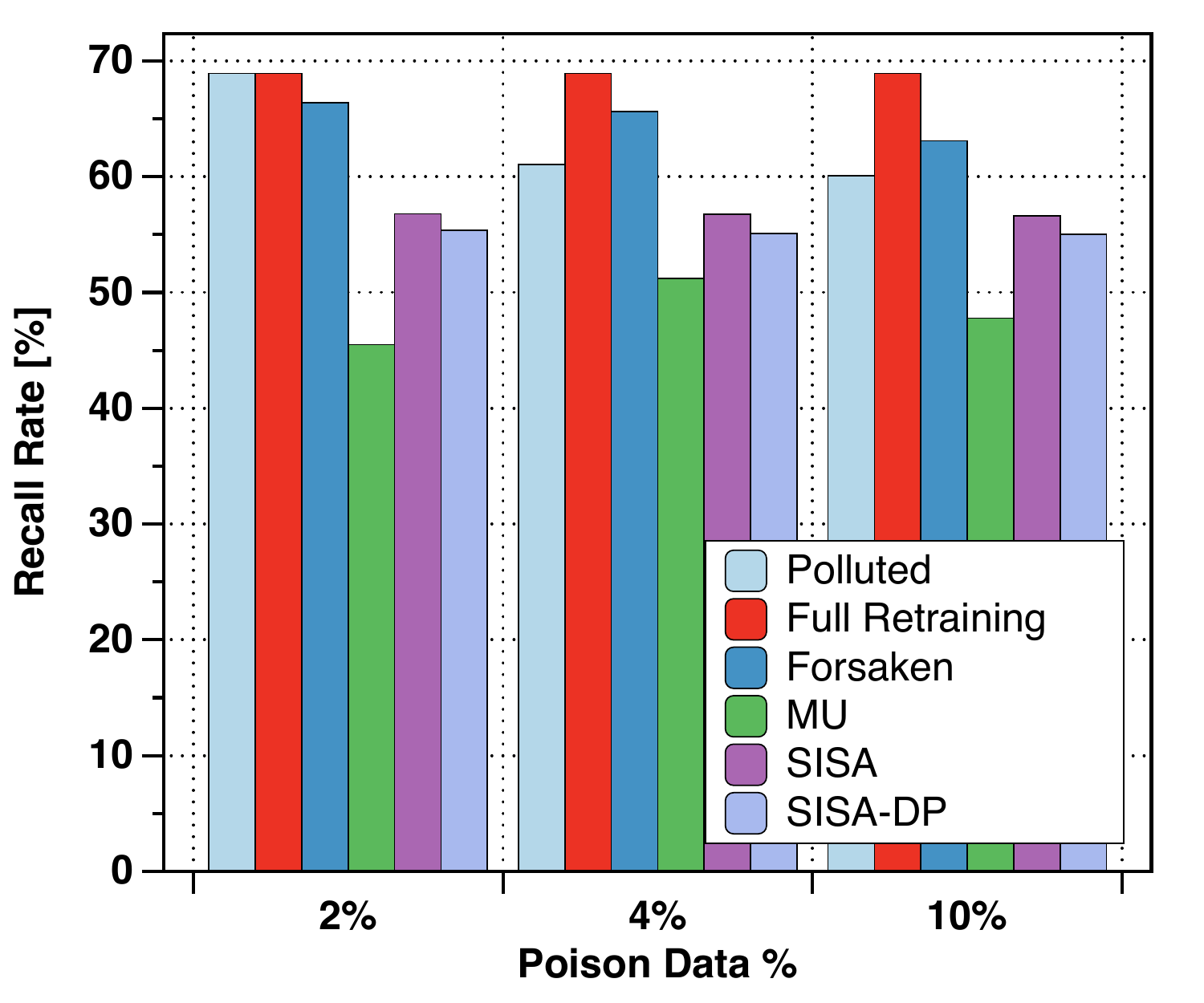}
    \label{fig_poison_data_recall}
    \end{minipage}
}%
\caption{\revision{The performance change of the target model under the poison data scenario.}}
\label{fig_poison_data}
\end{figure}

\subsection{Unintended Memorization Exposure}\label{sub_exposure}
For the generative sequence model, e.g, the LSTM~\cite{sundermeyer2012lstm} based language model, Carlini \et~\cite{carlini2019secret} introduced an indicator to measure the degree of the unintended memorization caused by a given OOD but private sequence $s[r]$, expressed as $exposure$.
As stated in~\cite{carlini2019secret}, the risk of the private information to be extracted by the adversary is positively related to $exposure$.
The phenomenon demonstrates the high data privacy risk caused by OOD data and becomes one of the key factors to make this paper focus more on OOD data unlearning.
The following is the equation to compute $exposure$.
\begin{equation}
    exposure_{\theta}(s[r]) = -\log_2 \int_{0}^{P_{\theta}(s[r])} \rho(x) dx,
\end{equation}
where $P_{\theta}(s[r])$ is the log-perplexity of $s[r]$ under the machine learning model $f_{\theta}$ and $\rho(\cdot)$ is a standard skew-normal function~\cite{o1976bayes} with mean $\mu$, standard derivation $\sigma^2$ and skew $\alpha$.
Log-perplexity is a standard indicator to evaluate how ``surprise'' the language model is to see a given sequence, which can be computed based on Eq.~\ref{eq_perplexity}.
\begin{equation}\label{eq_perplexity}
    P_{\theta}(s[r]) = \sum_{i = 1}^{n} (-\log_2 Pr(x_i| f_{\theta}(x_1 ... x_{i - 1}))).
\end{equation}
Notably, $exposure$ is a specifically defined indicator that can only be applied to the generative sequence model.

\begin{figure}[ht!]
\centering
\includegraphics[scale=0.35]{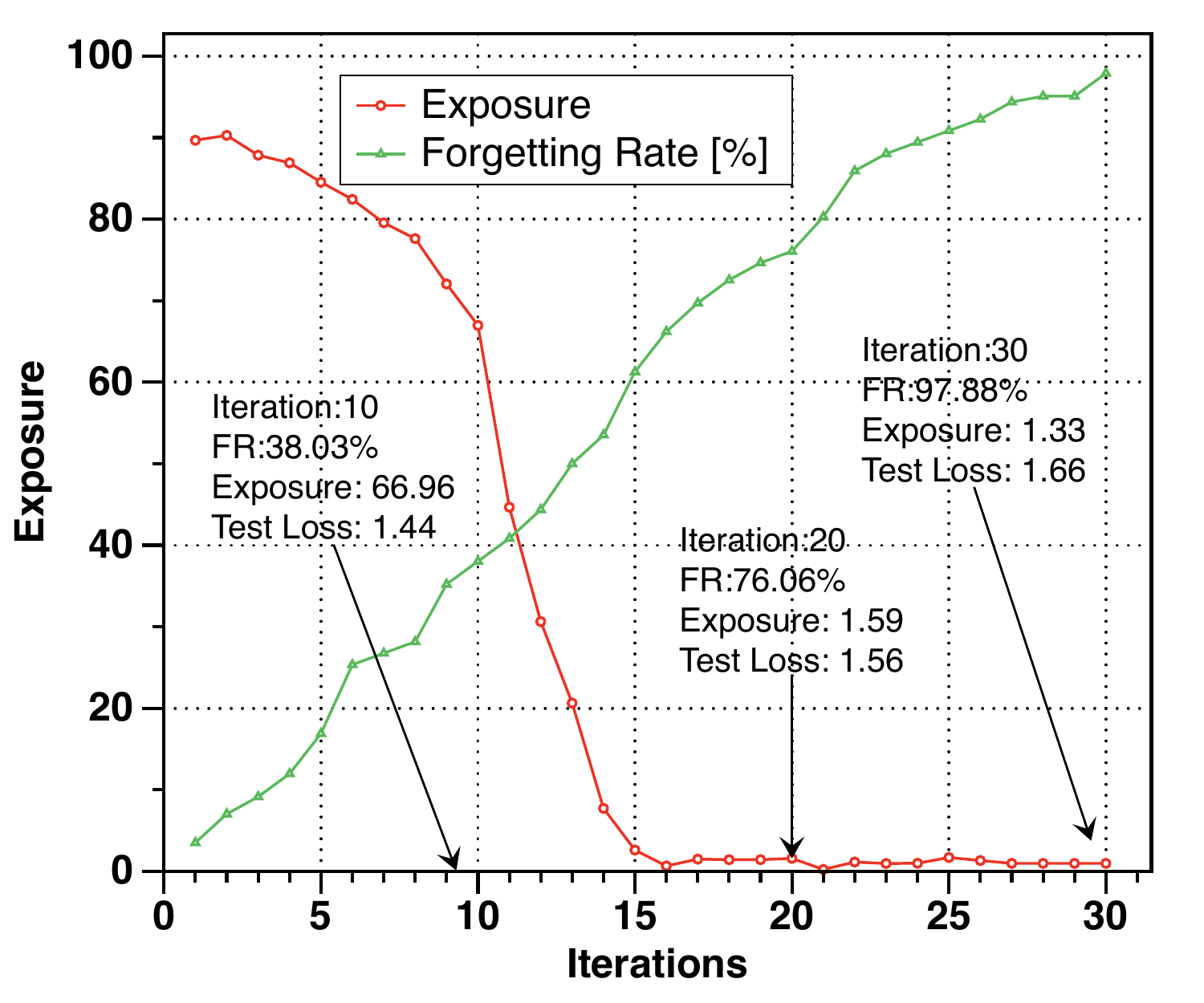}
\caption{Comparing $FR$ and test loss to $exposure$ across different training iterations of the mask gradient generator.}
\label{fig_language_elimination}
\end{figure}

To evaluate the effectiveness of \sysname to lower the risk of data privacy leakage, we use the Penn TreeBank (PTB) dataset~\cite{marcus1993building} to train a language model with a two-layer LSTM.
During training, we insert 100 canaries\footnote{Canary is a kind of manually generated sequence defined in~\cite{carlini2019secret}.} into the training set using the same method as~\cite{carlini2019secret}.
The 100 canaries are treated as the OOD samples required to be unlearned.
Fig.~\ref{fig_language_elimination} plots the experiment result, where we record the averaged $exposure$ of the 100 canaries.
It can be discovered that $exposure$ decreases along with the increase of $FR$.
According to~\cite{carlini2019secret}, the success probability of the adversary to extract a given sequence is negligible when $exposure$ is less than 10.
Therefore, \sysname can significantly lower the probability of the adversary to extract the user's private information from the unintended memorization.

\subsection{\revision{Expansion to Other Models}}
\label{sec_appendix_other_models}
\revision{
    \textlabel{Here, except for neural networks(NN) used in}{rC2:1} the previous experiments, we expand \sysname to other types of models, including logistic regression (LR) and support vector machine (SVM).
    Since LR and SVM are usually unable to achieve satisfying performance on image datasets, the experiments about image datasets are omitted.
    Clearly, based on Table~\ref{table_expansion}, \sysname can still achieve satisfying performance on both $FR$ and accuracy drops for these models, which further shows the robustness and stability of \sysname while applying to adaptive models.
}

\begin{table}[!htbp]
\centering
\footnotesize
\caption{\revision{Expansion of \sysname to other types of models}}

\resizebox{0.43\textwidth}{!}{%
\begin{tabular}{c|c|c|c|c}

\toprule
\multicolumn{1}{c|}{\textbf{Dataset}} & Indicator & \textbf{NN} & \textbf{LR} & \textbf{SVM} \\

\hline
\multirow{2}{*}{I.C.} & FR & 98.46\% & 98.53\% & 92.01\% \\

\cline{2-5}
& Diff.Test.Acc & 3.34\% & 4.82\% & 5.24\% \\

\hline
\multirow{2}{*}{Reuters (35-11)} & FR & 96.35\% & 96.11\% & 93.33\% \\

\cline{2-5}
& Diff.Test.Acc & 2.66\% & 4.49\% & 3.43\% \\

\hline
\multirow{2}{*}{News (15-5)} & FR & 97.53\% & 98.77\% & 90.85\% \\

\cline{2-5}
& Diff.Test.Acc & 4.61\% & 5.54\% & 6.09\% \\

\bottomrule
\end{tabular}
}
\label{table_expansion}
\end{table}





\section{Related Work}
\label{sec_relatedwork}
In this section, we briefly review the related works that inspire our design of \sysname.

\vspace{0.1cm}
\noindent
\textbf{Membership Oracle.}
The research of membership oracle is first inspired by the membership privacy problem while deploying machine learning as a service~\cite{shokri2017membership}.
Membership oracle answers the question of whether a given sample is a member of the training set used to train the target model based on the model output or not.
The principle of membership oracle inspires us a lot to define the evaluation indicator of machine unlearning for machine learning.
The early method to train a membership oracle used multiple shadow models~\cite{shokri2017membership}, which was complex and inefficient. 
Later, Salem \et~\cite{salem2018ml} showed that the membership oracle trained with only one shadow model could still have a good performance for most machine learning models.
Further work~\cite{truex2018towards} studied why the implementation of a membership oracle is possible.
To defend the attack launched by a membership oracle, Jia \et~\cite{jia2019memguard} believed that the most effective way was adding differential noises to the model predictions.

\vspace{0.1cm}
\noindent
\textbf{Data Forgetting of Machine Learning.}
Directive data forgetting has been one of the hard challenges for the security of machine learning~\cite{cao2015towards, bourtoule2019machine, ginart2019making, guo2019certified}.
Cao \et~\cite{cao2015towards} extensively studied the possibility of deploying the data forgetting mechanism in some real-world machine learning systems.
Similar to~\cite{ginart2019making} and~\cite{guo2019certified}, the main research object of~\cite{cao2015towards} was the machine learning models with a simple structure, e.g., Na\"{i}ve Bayes.
Bourtoule \et~\cite{bourtoule2019machine} extended data forgetting to neural networks by introducing a model retraining mechanism combined with ensemble learning.
For retraining, the first unavoidable problem is the high extra computation cost.
Although \cite{bourtoule2019machine} solves the problem to some extent by using the ensemble learning technique, it also dramatically change the training procedure of the original machine learning.
Also, there are some excellent work about the memorization of machine learning that are instructive for the research of data forgetting.
The model stealing of \cite{tramer2016stealing} first showed the possibility of extracting the learned memorization, i.e., trainable parameters, from a remote model, so that the adversary can copy a similar model by observing the predictions of the target model.
Then, Carlini \et~\cite{carlini2019secret} studied unintended memorization in the generative sequence model and gave an efficient method to measure the risk of extracting private data from the unintended memorization.
Besides the applicable models, the critical difference of~\cite{carlini2019secret} and \sysname is that the former focuses on how to extract the memorization of machine learning, but \sysname tries to eliminate the unexpected memorization.
\revision{
    \textlabel{Moreover, considering the indicator to}{rB2:1} evaluate machine unlearning, \cite{feldman2020neural} and \cite{jiang2020characterizing} proposed two related promising indicators, called influence estimate and C-score, respectively.
    These two indicators could quantitatively estimate the relationship between the OOD data or testing data and the normal training data.
    However, limited by their high computation complexity and inability to directly identify whether a given instance is forgotten or not, they are not included in the experiments of this paper.
}

\vspace{0.1cm}
\noindent
\textbf{Private Learning.}
A related idea for protecting the user data is based on cryptography or federated learning~\cite{bonawitz2017practical, tran2019federated, liu2017oblivious, mohassel2018aby3, agrawal2019quotient, nasr2018machine}.
The frequently used crytographic tools to ensure data confidential and non-identifiable in the schemes include oblivious transfer~\cite{liu2017oblivious}, differential privacy~\cite{geyer2017differentially}, homomorphic encryption~\cite{mohassel2018aby3}, secret sharing~\cite{bonawitz2017practical} or hybrid scheme~\cite{xu2019hybridalpha}.
Especially, for federated learning, its protection target is the transmission of gradients.
Therefore, although the objective of private learning is different from machine unlearning, it provides a powerful tool to enhance the data security of \sysname.
A typical example is that if we apply the privacy-preserving federated learning technique~\cite{bonawitz2017practical, smith2017federated, tran2019federated} to \sysname and upload the dummy gradient in the encrypted format, the attack surface of the adversary shall be theoretically limited to the physical device level.
The commonly accepted fact is that the hardness to hack a physical device is much higher than the communication channel.
The main defect that burdens the wider application of private learning is its hardness to balance the degree of privacy protection and additional computational overloads or data utility.

\vspace{-0.2cm}

\section{Conclusion} 
\label{Conclusion}
In this paper, we emphatically discussed the machine unlearning problem,
and inspired by the ``active forgetting'' mechanism of the human nervous system, proposed our ``learn to forget'' method, \sysname, to achieve machine unlearning for neural networks.
In the method, the user could stimulate the neurons of a machine learning model to unlearn specific memorization by training a dummy gradient generator.
In particular, the mask gradient could be treated in the same way as the common procedure of machine learning, which was more practical and efficient than the prior works.
For better evaluating the performance of machine unlearning, we also presented the first indicator, called forgetting rate, to measure the transformation rate of the eliminated data from ``memorized'' to ``unknown''.
\revision{
    Reconsidering the design of \sysname or other machine unlearning methods, \textlabel{we failed to provide provable guarantee}{rE5:1} for hiding the indirect footprint of unlearned data points.
    With our work as the stepping stone, we hoped more future works could further dive into the machine unlearning field and provide more extensive options to protect both private user data security and data unlearning privacy.
}

%
%

\normalsize
\bibliographystyle{IEEEtran}
\bibliography{references}
%





\appendix
\section*{Appendices}
\addcontentsline{toc}{section}{Appendices}
\renewcommand{\thesubsection}{\Alph{subsection}}

\subsection{Forgetting Rate}
\label{sec_appendix_fr}
In Section~\ref{sec_approach}, we define the indicator $FR$ to evaluate the performance of machine unlearning.
Here, we outline two scenarios to explain why $FR$ can evaluate the performance of machine unlearning.

First, we consider an ideal scenario.
Assume both the membership oracle and the machine unlearning method are ideal, and the size of unlearned OOD data is $100$.
In such a condition, the membership oracle can $100\%$ identify whether a given sample is used for training.
Thus, we have $BT = 100$.
Further, since the machine unlearning method is perfect, all unlearned data become ``unknown'' after machine unlearning, which means $AF = 100$.
At this time, $FR$ reaches its upper bound, $FR = \frac{100 - 0}{100} \times 100\% = 100\%$.
Similarly, we can prove $FR$ can achieve its lower bound when $AF = 0$ and none of OOD data are eliminated.

Turn to a more practical scenario.
Assume that there is a user who wants to unlearn the memorization of an image classification model about his $100$ images.
As the inspector, we observe that before memorization, $BF = 10$ unlearned samples have been labeled as false by the membership oracle.
After machine unlearning, $AF = 90$ unlearned samples are identified as false.
According to Eq.~\ref{eq_forgetting_rate}, we have to first remove the $10$ samples that originally unknown, i.e., $AF - BF = 90 - 10$.
Then, it can be computed that $FR = \frac{90 - 10}{90} = 0.89$.
Here, $FR$ indicates that $89\%$ unlearned data is successfully unlearned due to the machine unlearning algorithm.

\subsection{Catastrophic Forgetting Rate}
\label{sec_appendix_eel}
\revision{
    \textlabel{$FR$ only considers the status of unlearned data but ignore the status of other data.}{cr3:1}
    Consider a scenario where all data points, not just the unlearned ones, are classified as false by the membership oracle after machine unlearning.
    In the scenario, the machine unlearning algorithm attains a good performance on $FR$, but the classification confidence of the target model over the remaining data suffers from a dramatic loss.
    To better evaluate machine unlearning algorithms, an auxiliary indicator called \textit{catastrophic forgetting rate} ($CFR$) is introduced.
    $CFR$ measures the percentage of normal training samples whose memorization is excessively unlearned by the machine unlearning algorithm.
    The computation of $CFR$ is also based on the membership oracle, and defined in Eq.~\mbox{\ref{eq_excessive}}.
}
\begin{equation}\label{eq_excessive}
\begin{gathered}
    $CFR$ = \frac{BT.Train - AT.Train}{BT.Train},
\end{gathered}
\end{equation}
\revision{
    where $BT.Train$ and $AT.Train$ represent the number of training data that are memorized before and after conducting memorization, respectively.
    A good machine unlearning algorithm should achieve a high $FR$ and low $CFR$.
    Different from accuracy drop (discussed in Section~\ref{sec_experiments}) that only considers hard-label change caused by machine unlearning, $CFR$ provides more elaborate evaluation from the posterior perspective.
    Experiments about $CFR$ are given below.
    Here, only OOD data unlearning results are demonstrated for brevity.
}

\renewcommand\arraystretch{1.2}
\begin{table}[!htbp]
\centering
\footnotesize
\caption{\revision{$CFR$s of \mbox{\sysname} on Different Datasets. A better machine unlearning algorithm tends to have higher $FR$ and lower $CFR$ as discussed in Section~\mbox{\ref{sec_approach}}.}}

\resizebox{0.47\textwidth}{!}{%
\begin{tabular}{
    c  |c | c | c | c
}

\toprule
\multirow{2}{*}{Dataset}  & \multicolumn{4}{c}{$CFR$}\\

\cline{2-5}
& \sysname & MU~\cite{cao2015towards} & SISA~\cite{bourtoule2019machine} & SISA-DP~\cite{bourtoule2019machine} \\

\hline
C10.S. &  13.42\% & 79.92\% & 23.91\% & 24.45\% \\

\hline
C10.T. &  14.87\% & 70.71\% & 21.52\%  & 28.63\% \\

\hline
C100.T. &  6.21\% & 72.33\% & 41.31\% & 36.65\% \\

\hline
I.C.    &  2.92\% & 5.9\% & 2.19\% & 4.54\%\\

\hline
Reuters (35-11) & 16.32\% & 32.77\% & 17.27\% & 28.23\% \\

\hline
News (15-5) &  19.92\% & 82.15\% & 26.3\% & 22.63\% \\

\bottomrule

\end{tabular}
}
\label{table_eel}
\end{table}

\revision{
    As shown in Table~\mbox{\ref{table_eel}}, \mbox{\sysname} achieves the lowest $CFR$ in all experiments.
    Not surprisingly, the catastrophic forgetting phenomenon of MU is the most obvious and its $CFR$ is the highest.
    Here, an extraordinary phenomenon is that the retraining based method, SISA and SISA-DP, does not achieve the best performance.
    The reason is that the partition of the training samples leads to the consequence that there are not enough samples for each constituent model to form strong memorization on all training data.
    In other words, the ensemble model in SISA cannot memorize all training data in some applications, especially when the learning task is complex and the size of the training set is not very large, like, C100.T.
    Furthermore, combined with Fig.~\ref{fig_test_acc}, we observe that $CFR$ has a positive relation to the increase of accuracy drops, which reflects the effectiveness of our proposed indicator.
}

\subsection{Dataset}
\label{sec_appendix_dataset}
In the experiments, we mainly utilize eight datasets to evaluate the effectiveness of \sysname, shown in Table~\ref{table_datasets}.
The datasets can be classified into three categories, which are image classification, sentiment analysis and text categorization.
Among them, CIFAR10\footnote{https://www.cs.toronto.edu/~kriz/cifar.html}, CIFAR100, STL-10 and TinyImage are used for image classification.
IMDB\footnote{https://ai.stanford.edu/~amaas/data/sentiment/} and Customer Review are sentiment analysis datasets.
The remaining two datasets, Reuster\footnote{provided by Kersa platform, https://keras.io/api/datasets/reuters/} and News\footnote{https://archive.ics.uci.edu/ml/datasets/Twenty+Newsgroups}, are for text categorization.
Especially, STL-10\footnote{https://ai.stanford.edu/~acoates/stl10/}, TinyImage\footnote{https://tiny-imagenet.herokuapp.com/} and Customer\footnote{https://github.com/hendrycks/error-detection/tree} are used to provide OOD data and form unintended memorization.
The performance of the trained neural networks for the datasets is listed in Table~\ref{table_neural_network_performance}.
In the Table, C.B.OOD means the number of OOD samples that are correctly identified before machine unlearning.

\begin{table}[!htbp]
\centering
\footnotesize
\caption{Dataset Information}
\begin{tabular}{|p{1.33cm}<{\centering}|c|c|p{0.73cm}<{\centering}|p{0.73cm}<{\centering}|}

\hline
Name & No. of Instances & Features & Classes & Epochs \\

\hline
CIFAR-10 & 60000 (10k for testing) & $32\times 32$ & 10 & 20 \\

\hline
CIFAR-100 & 60000 (10k for testing) & $32\times 32$ & 100 & 15 \\

\hline
IMDB & 50000 (25k for testing) & 200 & 2 & 30 \\

\hline
Reuters & 11228 (2246 for testing) & 10000 & 46 & 40 \\

\hline
News & 11314 (2262 for testing) & 1000 & 20 & 40 \\

\hline
STL10 & 13000 & $32\times 32$ & 10 & OOD \\

\hline
TinyImage & 100000 & $32\times 32$ & 200 & OOD \\

\hline
Customer & 2775 & 200 & 2 & OOD \\

\hline
\end{tabular}
\label{table_datasets}
\end{table}

\begin{table}[!htbp]
\centering
\footnotesize
\caption{Neural Network Information}
\resizebox{0.47\textwidth}{!}{%
\begin{tabular}{c|c|c|c|c|c}

\toprule
\multirow{2}{*}{\textbf{Dataset}} & \multicolumn{2}{c|}{\textbf{Acc with OOD Data}} & \multicolumn{3}{c}{\textbf{Acc without OOD data}} \\

\cline{2-6}
 & \textbf{Acc.Train} & \textbf{Acc.Test} & \textbf{Acc.Train} & \textbf{Acc.Test} & \textbf{C.B.OOD} \\

\hline
C10.S. & 96.15\% & 72.72\% & 99.99\% & 77.98\% & 120 \\

\hline
C10.T. & 99.99\% & 76.92\% & 99.99\% & 77.98\% & 181  \\

\hline
C100.T. & 99.9\% & 47.89\% & 99.99\% & 50.74\% & 155 \\

\hline
IMDB & 90.71\% & 85.34\% & 91.68\% & 85.7\% & 185 \\

\hline
Reuters & 90.62\%  & 77.89\% & 91.96\% & 79.91\% & 146 \\

\hline
News & 99.93\% & 61.57\% & 99.99\% & 64.93\% & 154 \\

\bottomrule

\end{tabular}
}
\label{table_neural_network_performance}
\end{table}

\subsection{Membership Inference}
\label{sec_appendix_membership_inference}
In the experiments, the performance of the membership oracle strongly affects the computation results of $FR$. 
Thus, we list the accuracy, precision and recall rate for the membership oracles trained for each dataset in Table~\ref{table_membership_inference} to make it more intuitive to evaluate our experiment results.
The record in Table~\ref{table_membership_inference} is the experiment result over the whole training and testing datasets, not only the unlearning data set.
\begin{table}[!htbp]
\centering
\footnotesize
\caption{The accuracy, precision and recall rate of the membership oracle for each dataset}

\resizebox{0.4\textwidth}{!}{%
\begin{tabular}{c|c|c|c}

\toprule
\multirow{2}{*}{\textbf{Dataset}} & \multicolumn{3}{c}{\textbf{Membership Oracle (\%)}}\\

\cline{2-4}
 & Accuracy & Precision & Recall Rate \\

\hline
C10.S.   & 83.48 & 82.17 & 87.74 \\

\hline
C10.T.   & 84.21 & 83.33 & 89.48 \\

\hline
C100.T.  & 95.88 & 95.50 & 99.76 \\

\hline
I.C.     & 75.49 & 75.31 & 88.38 \\

\hline
Reuters (35-11) & 80.18 & 82.01 & 86.62 \\

\hline
News (15-5)     & 86.04 & 86.64 & 87.59 \\

\bottomrule
\end{tabular}
}
\label{table_membership_inference}
\end{table}



\subsection{Neural Networks}
\label{sec_appendix_network}
For CIFAR-10 and CIFAR100, we use a previously proposed deep network architecture VGG-16~\cite{simonyan2014very}.
Fig.~\ref{fig_network_IMDB}, Fig.~\ref{fig_network_Reuters} and Fig.~\ref{fig_network_news} show the neural network architectures used to process IMDB, Reuters and News, respectively.

\begin{figure}[h]
	\centering
	\begin{boxedminipage}{0.46\textwidth}
        \footnotesize
		\begin{enumerate}
		    \item \textit{Embedding:} Input word $1\times 200$, the output word embedding is $1\times 100$.
		
			\item \textit{Convolution: } Windows size $3\times 100$, number of output channel $100$.
			
			\item \textit{Convolution: } Windows size $4\times 100$, number of output channel $100$.
			
			\item \textit{Convolution: } Windows size $5\times 100$, number of output channel $100$.
			
			\item \textit{MaxPooling:} Window Size $1\times 35\times 35$.
			
			\item \textit{Fully Connected Layer:} Fully connected the incoming $300$ nodes to the outgoing $2$ nodes.

		\end{enumerate}
		\normalsize
	\end{boxedminipage}
	\caption{The neural network used for IMDB.}
	\label{fig_network_IMDB}
\end{figure}

\begin{figure}[h]
	\centering
	\begin{boxedminipage}{0.46\textwidth}
        \footnotesize
		\begin{enumerate}
			\item \textit{Fully Connected Layer:} Fully connected the incoming $10000$ nodes to the outgoing $512$ nodes.
			
			\item \textit{ReLU:} Calculate ReLU for each input.
			
			\item \textit{Dropout:} Dropout rate is $0.5$.
			
			\item \textit{Fully Connected Layer:} Fully connected the incoming $512$ nodes to the outgoing $256$ nodes.
			
			\item \textit{ReLU:} Calculate ReLU for each input.
			
			\item \textit{Dropout:} Dropout rate is $0.5$.
			
			\item \textit{Fully Connected Layer:} Fully connected the incoming $256$ nodes to the outgoing $35$ nodes.

		\end{enumerate}
		\normalsize
	\end{boxedminipage}
	\caption{The neural network used for Reuters.}
	\label{fig_network_Reuters}
\end{figure}

\begin{figure}[h]
	\centering
	\begin{boxedminipage}{0.46\textwidth}
        \footnotesize
		\begin{enumerate}
		    \item \textit{Embedding:} Input word $1\times 1000$, the output word embedding is $1\times 100$.
		
			\item \textit{Transpose Convolution: } Windows size $5\times 5$, output channel $128$.
			
			\item \textit{ReLU:} Calculate ReLU for each input.
			
			\item \textit{MaxPooling:} Window Size $1\times 5\times 5$.
			
			\item \textit{Convolution: } Windows size $5\times 5$, number of output channel $128$.
			
			\item \textit{ReLU:} Calculate ReLU for each input.
			
			\item \textit{MaxPooling:} Window Size $1\times 5\times 5$.
			
			\item \textit{Convolution: } Windows size $5\times 5$, number of output channel $256$.
			
			\item \textit{ReLU:} Calculate ReLU for each input.
			
			\item \textit{MaxPooling:} Window Size $1\times 35\times 35$.
			
			\item \textit{Fully Connected Layer:} Fully connected the incoming $128$ nodes to the outgoing $128$ nodes.
			
			\item \textit{Fully Connected Layer:} Fully connected the incoming $128$ nodes to the outgoing $20$ nodes.
		\end{enumerate}
		\normalsize
	\end{boxedminipage}
	\caption{The neural network used for News.}
	\label{fig_network_news}
\end{figure}
\vspace{-0.1cm}

\subsection{\revision{Big Dataset}}
\label{sec_appendix_big_dataset}
\revision{
    \textlabel{To further evaluate the improvement of}{rB3:2} \sysname on efficiency for machine unlearning, we also try some experiments about TinyImage, a larger dataset than the currently used ones.
    In the experiments, we randomly select 90 categories of TinyImage to form the target dataset and the remaining categories to form the OOD dataset.
    The unlearning size is 200.
    The used neural network is VGG-16.
    As shown in Table~\ref{table_big_dataset}, \sysname achieves similar performance to SISA and SISA-DP on $FR$.
    However, since the complexity of \sysname is not related to the target dataset size, the running time of \sysname with TinyImage is almost the same as the one in CIFAR10 and much faster than other methods.
}

\begin{table}[!htbp]
\centering
\footnotesize
\caption{\revision{Comparison with TinyImage.}}
\resizebox{0.47\textwidth}{!}{%
\begin{tabular}{c|c|c|c|c|c}

\toprule
\centering \textbf{Dataset}
& \textbf{Retraining}
& \textbf{\sysname}
& \centering \textbf{MU} 
& \textbf{SISA}
& \textbf{SISA-DP} \\


\hline
$FR$          & 93.51\% & 96.63\% & 75.7\% & 91.59\% & 93.61\%  \\

\hline
Test.Acc.Diff & +2.06\% & -3.12\% & -47.05\% & -12.25\% & -14.41\%  \\

\hline
Running Time  & 2011.31 & 16.09 & 54.12 & 1923.59 & 1925.85 \\

\bottomrule
\end{tabular}
}
\label{table_big_dataset}
\end{table}

\subsection{\revision{ID Data Machine Unlearning with Different Unlearning Sizes}}\label{sec_appendix_unlearning_id}
\revision{
    Fig.~\mbox{\ref{fig_eliminated_size_id}} illustrates the FR and accuracy drop comparison results with different unlearning sizes for ID data unlearning.
    The experimental results show that even more accuracy drops are introduced for ID data unlearning, \sysname still outperforms other methods and attains the best performance.
}

\begin{figure}[htbp]
\centering
\subfigure[The change of $FR$ with different unlearning sizes for CIFAR10 for in-distribution data.]{
\begin{minipage}[t]{0.45\linewidth}
\centering
\includegraphics[scale=0.24]{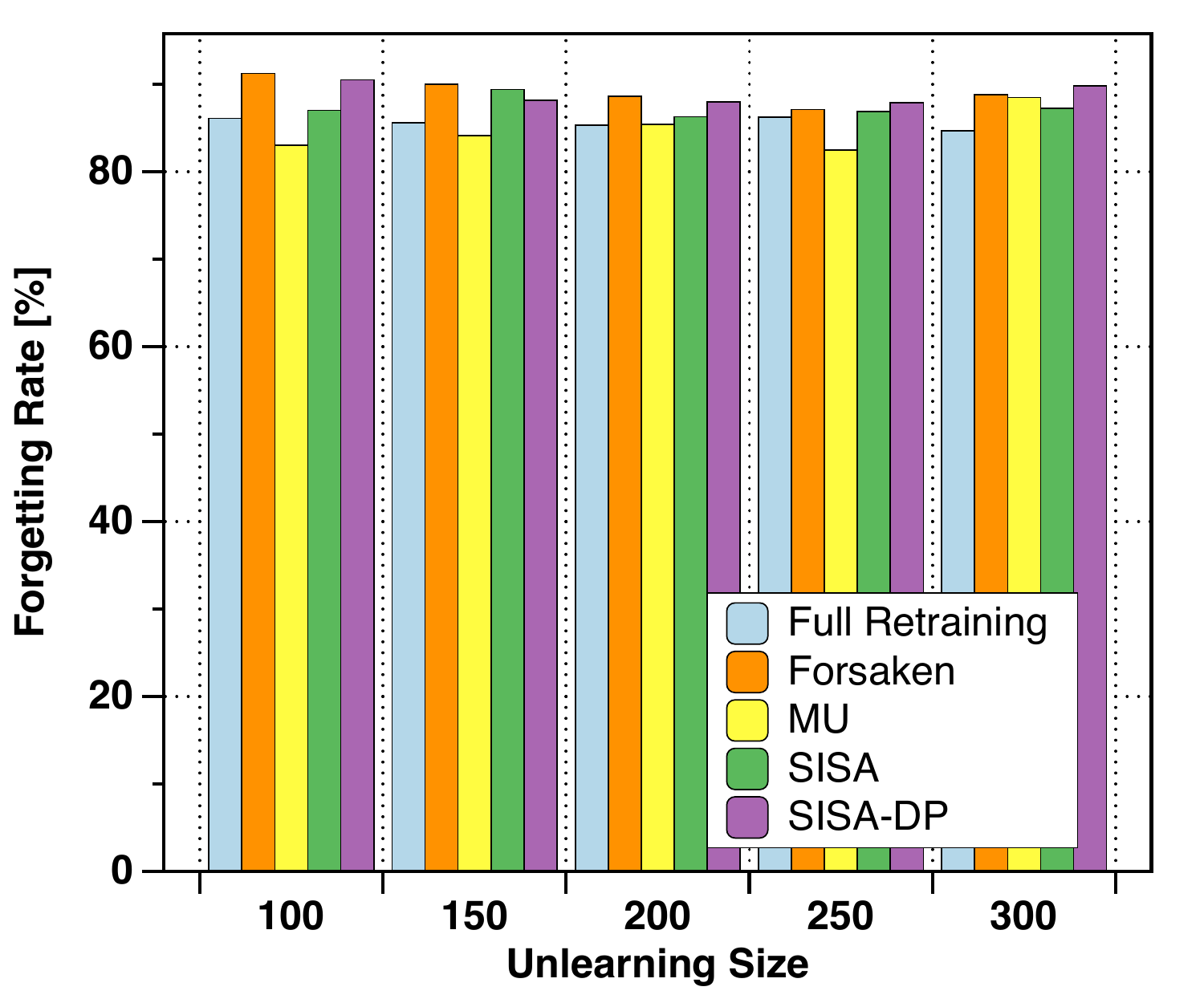}
\label{fig_eliminated_size_cifar_acc_id}
\end{minipage}
}%
\hfill
\subfigure[The change of Test.Acc with different unlearning sizes for CIFAR10 for in-distribution data.]{
\begin{minipage}[t]{0.45\linewidth}
\centering
\includegraphics[scale=0.24]{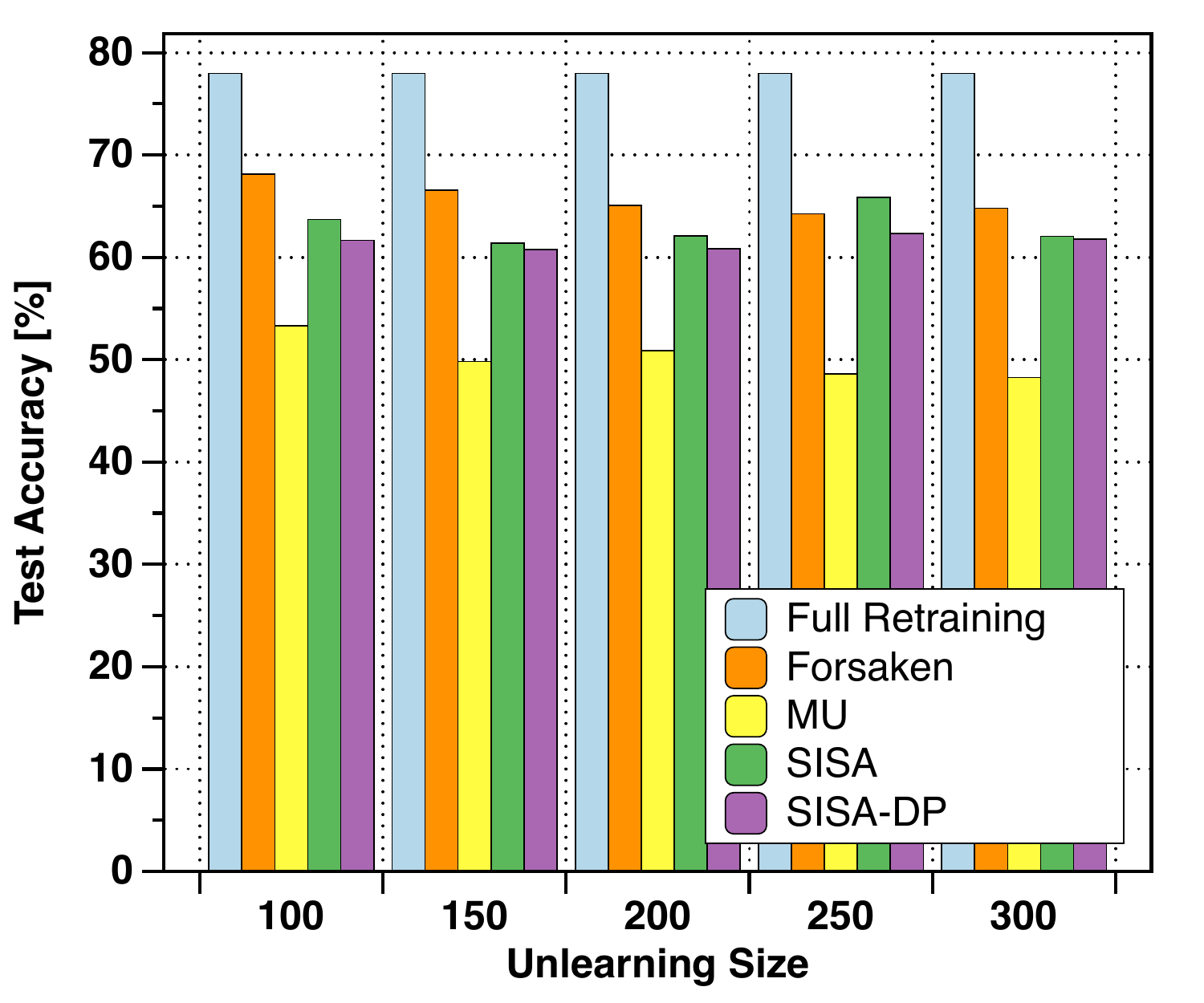}
\label{fig_eliminated_size_cifar_fr_id}
\end{minipage}
}%
\hfill
\subfigure[The change of $FR$ with different unlearning sizes for IMDB for in-distribution data.]{
\begin{minipage}[t]{0.45\linewidth}
\centering
\includegraphics[scale=0.24]{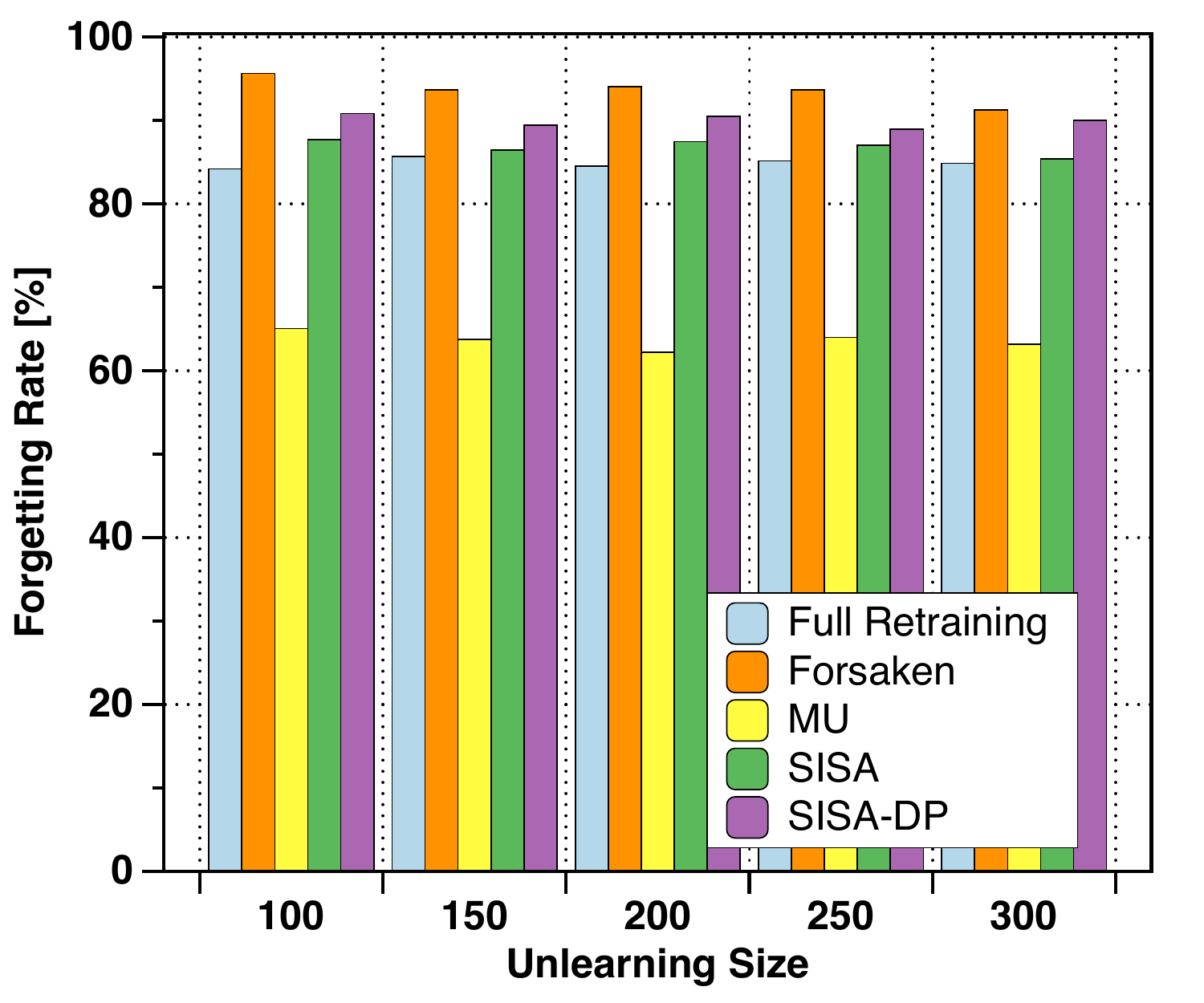}
\label{fig_eliminated_size_imdb_acc_id}
\end{minipage}
}%
\hfill
\subfigure[The change of Test.Acc with different unlearning sizes for IMDB for in-distribution data.]{
\begin{minipage}[t]{0.45\linewidth}
\centering
\includegraphics[scale=0.24]{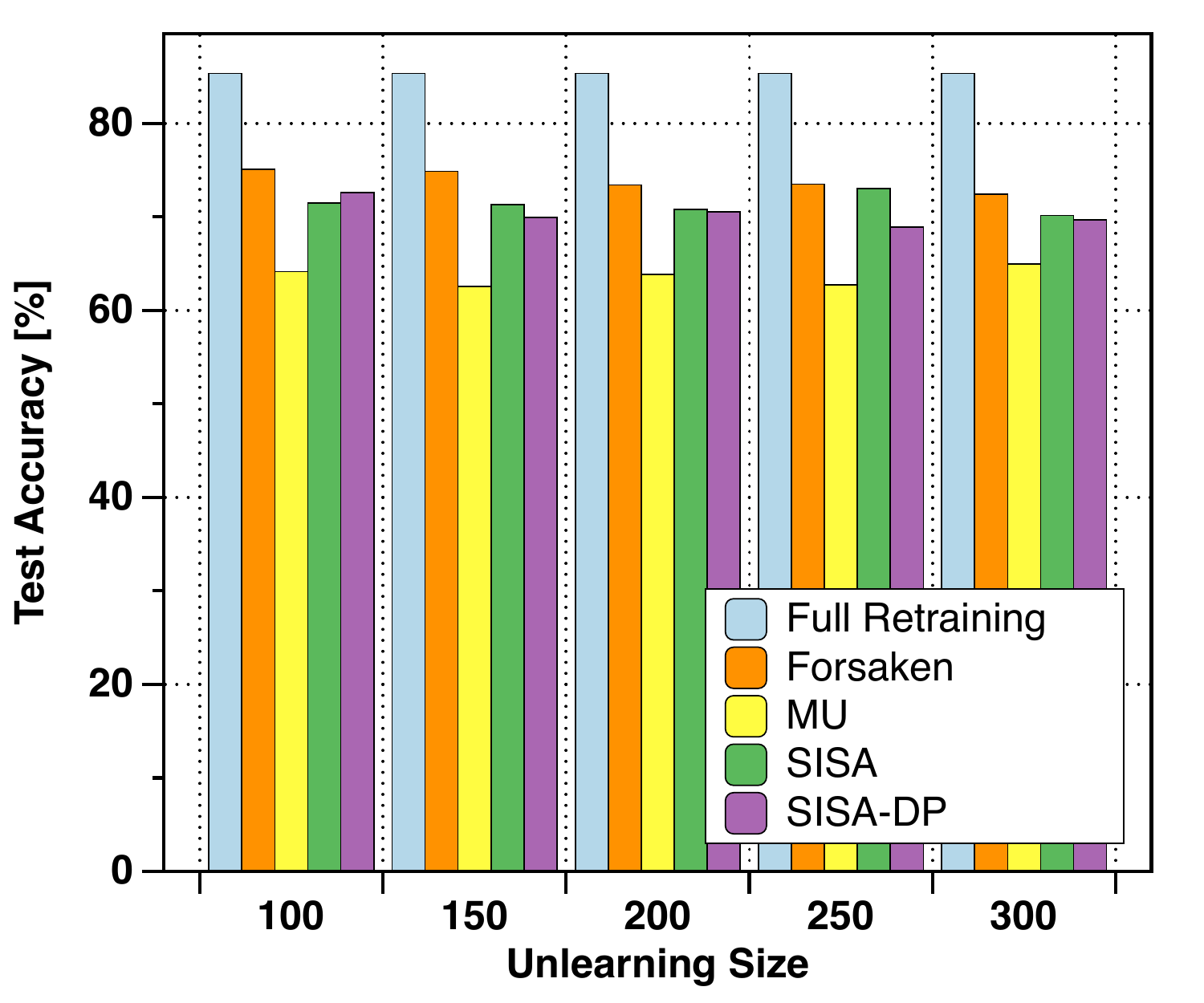}
\label{fig_eliminated_size_imdb_fr_id}
\end{minipage}
}%
\caption{
\revision{The performance change comparison with different unlearning sizes for ID data unlearning.
}
}
\label{fig_eliminated_size_id}
\end{figure}

\subsection{\revision{Efficiency Analysis}}\label{sec_appendix_running_time}
\revision{
    In Table~\mbox{\ref{table_efficiency_eliminated_size}}, we show the running time changes of \mbox{\sysname} with different unlearned data sizes.
    For each kind of learning task, we select a dataset.
    From the results, the running time \mbox{\sysname} almost increases linearly.
    Compared with other baselines shown in Table~\mbox{\ref{table_efficiency_comparison}}, \sysname is still faster for tens of times even with 300 samples (more than 1\%) required to be forgotten.
}

\begin{table}[!h]
\centering
\footnotesize
\caption{\revision{Running time of machine unlearning with different unlearning sizes}}
\begin{tabular}{c|c|c|c|c|c|c}

\toprule
\multirow{2}{*}{Enlearning Size} & \multicolumn{6}{c}{Running Time (s)} \\

\cline{2-7}
& 50 & 100 & 150 & 200 & 250 & 300\\

\hline
C10.T.   & 4.12 & 8.01 & 13.71 & 16.12 & 20.44 & 25.09 \\

\hline
I.C.     & 0.98 & 1.39 & 2.08 & 2.35 & 3.11 & 3.84 \\

\hline
Reuters (35-11)  & 0.55 & 1.01 & 1.48 & 1.68 & 2.21 & 3.18 \\

\bottomrule
\end{tabular}
\label{table_efficiency_eliminated_size}
\end{table}

\end{document}